\PassOptionsToPackage{table,xcdraw}{xcolor}
\documentclass[nohyperref]{article}

\usepackage[preprint]{template/neurips_2023}

\usepackage[titletoc]{appendix}
\usepackage{amsmath,amsbsy,amssymb,amsfonts,amsthm}
\usepackage{mathtools}
\usepackage{color}
\usepackage{xspace} 
\usepackage{times}
\usepackage{graphicx}
\usepackage{subcaption}
\usepackage{algorithm,algorithmic}
\usepackage{tcolorbox}
\usepackage{wrapfig}

\usepackage[utf8]{inputenc} 
\usepackage[T1]{fontenc}    
\definecolor{mydarkblue}{rgb}{0,0.08,0.45}
\usepackage[colorlinks=true,
    linkcolor=mydarkblue,
    citecolor=mydarkblue,
    filecolor=mydarkblue,
    urlcolor=mydarkblue]{hyperref}      
\usepackage{url}            
\usepackage{booktabs,tabularx}     
\usepackage{nicefrac}      
\usepackage{microtype}     
\newcolumntype{Y}{>{\raggedright\arraybackslash}X} 
\newcolumntype{C}{>{$}c<{$}} 

\newcommand{\absval}[1]{\lvert#1\rvert}

\DeclareMathOperator*{\Var}{Var}

\title{Towards Out-of-Distribution Adversarial Robustness}

\author{Adam Ibrahim \\
Mila \\Université de Montréal \\
\texttt{first.last@mila.quebec}
\And
Charles Guille-Escuret \\
Mila \\Université de Montréal
\And
Ioannis Mitliagkas \\
Mila \\Université de Montréal
\And
Irina Rish \\
Mila \\Université de Montréal
\And
David Krueger \\
University of Cambridge
\And
Pouya Bashivan \\
Mila \\McGill University
}

\begin{document}
\maketitle
\begin{abstract}
Adversarial robustness continues to be a major challenge for deep learning. A core issue is that robustness to one type of attack often fails to transfer to other attacks. While prior work establishes a theoretical trade-off in robustness against different $L_p$ norms, we show that there is potential for improvement against many commonly used attacks by adopting a domain generalisation approach.
Concretely, we treat each type of attack as a domain, and apply the Risk Extrapolation method (REx), which promotes similar levels of robustness against all training attacks. Compared to existing methods, we obtain similar or superior worst-case adversarial robustness on attacks seen during training. Moreover, we achieve superior performance on families or tunings of attacks only encountered at test time. On ensembles of attacks, our approach improves the accuracy from 3.4\% with the best existing baseline to 25.9\% on MNIST, and from 16.9\% to 23.5\% on CIFAR10.

%
%
\end{abstract}

\section{Introduction}

Vulnerability to adversarial perturbations~\citep{biggio2013evasion, szegedy2014intriguing, goodfellow2015explaining} is a major concern for real-world applications of machine learning such as healthcare~\citep{qayyum2020secure} and autonomous driving~\citep{deng2020analysis}. 
For example, \citet{eykholt2018robust} show how seemingly minor physical modifications to road signs may lead autonomous cars into misinterpreting stop signs, while \citet{li2020practical} achieve high success rates with over-the-air adversarial attacks on speaker systems.

Much work has been done on defending against adversarial attacks~\citep{goodfellow2015explaining, papernot2016distillation}. However, new attacks commonly overcome existing defenses~\citep{athalye2018obfuscated}. A defense that has so far passed the test of time against individual attacks is adversarial training. \citet{goodfellow2015explaining} originally proposed training on examples perturbed with the Fast Gradient Sign Method (FGSM), which performs a step of sign gradient ascent on a sample $x$ to increase the chances of the model misclassifying it. \citet{madry2018towards} further improved robustness by training on Projected Gradient Descent (PGD) adversaries, which perform multiple updates of (projected) gradient ascent to try to generate a maximally confusing perturbation within some $L_p$ ball of predetermined radius $\epsilon$ centred at the chosen data sample. 

Unfortunately, adversarial training can fail to provide high robustness against several attacks, or tunings of attacks, only encountered at test time. For instance, simply changing the norm constraining the search for adversarial examples with PGD has been shown theoretically and empirically~\citep{khoury2018geometry, tramer2019adversarial, maini2020adversarial} to induce significant trade-offs in performance against PGD of different norms. This issue highlights the importance of having a well-defined notion of ``robustness'': while using the accuracy against individual attacks has often been used as a proxy for robustness, a better notion of robustness, as argued by \citet{athalye2018obfuscated}, is to consider the accuracy against an ensemble of attacks within a threat model (i.e.\ a predefined set of allowed attacks). Indeed, in the example of autonomous driving, an attacker will not be constrained to a single attack on stop signs, and is free to attempt several attacks to find one that succeeds.

In order to be robust against multiple attacks, we draw inspiration from domain generalisation. In domain generalisation, we seek to achieve consistent performance even in case of unknown distributional shifts in the inputs at test time. We interpret different attacks as distinct distributional shifts in the data, and propose to leverage existing techniques from the out-of-distribution generalisation literature.

We choose variance REx~\citep{krueger2021out}, which consists in using as a loss penalty the variance on the different training domains of the empirical risk minimisation loss. We choose this method as it is conceptually simple, its iterations are no more costly than existing multi-perturbation baselines', it does not constrain the architecture, and it can be used on models pretrained with existing defenses. We consider robustness against an adversary having access to both the model and multiple attacks. 

However, there are multiple potential challenges: first, \citet{gulrajani2020search} show that domain generalisation methods, such as REx, often fail to improve over empirical risk minimisation (ERM) in many settings. Thus, it is possible that REx would fail to improve \citet{tramer2019adversarial}'s defense, which uses ERM. Second, domain generalisation methods are usually designed for stationary settings, whereas in adversarial machine learning, the distribution of adversarial perturbations is non-stationary during training as the attacks adapt to the changes in the model parameters. Finally, the state-of-the-art multi-perturbation defense proposed by \citet{maini2020adversarial}, which we intend to improve with REx, does not explicitly train on multiple domains, which REx originally requires. 

Therefore, we are interested in the two following research questions:
\begin{enumerate}
    \item Can REx improve robustness against multiple attacks seen during training?
    \item Can REx improve robustness against unseen attacks, that is, attacks only seen at test time?
\end{enumerate}
Our results show that the answer to both questions is yes on the ensembles of attacks used in this work. We show that REx consistently yields benefits across variations in: datasets, architectures, multi-perturbation defenses, hyperparameter tuning, attacks seen during training, and attack types or tunings only encountered at test time. 

\vspace{-2mm}
\section{Related Work}
\vspace{-2mm}
\subsection{Adversarial attacks and defenses} 
\vspace{-2mm}
Since the discovery of adversarial examples against neural networks~\citep{szegedy2014intriguing}, numerous approaches for finding adversarial perturbations (i.e. adversarial attacks) have been proposed~\citep{goodfellow2015explaining,madry2018towards,moosavi2016deepfool,carlini2017towards,croce2020reliable}, with the common goal of finding perturbation vectors with constrained magnitude that, when added to the network's input, lead to (often highly confident) misclassification.

One of the earliest attacks, the Fast Gradient Sign Method (FGSM)~\citep{goodfellow2015explaining}, computes a perturbation on an input $x^0$ by performing a step of sign gradient ascent in the direction that increases the loss $L$ the most, given the model's current parameters $\theta$. This yields an adversarial example $\Tilde{x}$ that may be misclassified:
\begin{align}
\Tilde{x} = x^0 + \alpha \operatorname{sgn}(\nabla_x L(\theta,x^0,y)).
\end{align}
This was later enhanced into the Projected Gradient Descent (PGD) attack~\citep{kurakin2017adversarial,madry2018towards} by iterating multiple times this operation and adding projections to constrain it to some neighbourhood of $x^0$, usually a ball of radius $\epsilon$ centered at $x^0$, noted $\mathcal{B}_\epsilon(x^0)$:
\begin{align}\label{eq:PGD}
x^{t+1} = \Pi_{\mathcal{B}_\epsilon(x^0)} \left( x^t + \alpha\operatorname{sgn}(\nabla_x L(\theta,x^t,y))\right).
\end{align}

With the advent of diverse algorithms to \emph{defend} classifiers against such attacks, approaches for discovering adversarial examples have become increasingly more complex over the years. Notably, it was found that a great number of adversarial defenses rely on \emph{gradient obfuscation}~\citep{athalye2018obfuscated}, which consists in learning how to mask or distort the classifier's gradients to prevent attacks iterating over gradients from making progress. However, it was later discovered that such approaches can be broken by other attacks~\citep{athalye2018obfuscated,croce2020reliable}, some of which bypass these defenses by not relying on gradients~\citep{brendel2019accurate,andriushchenko2020square}. 

A defense that was shown to be robust to such countermeasures is Adversarial Training~\citep{madry2018towards}, which consists in training on adversarial examples. Adversarial training corresponds to solving a minimax optimisation problem where the inner loop executes an adversarial attack algorithm, usually PGD, to find pertubations to the inputs that maximise the classification loss, while the outer loop tunes the network parameters to minimise the loss on the adversarial examples. Despite the method's simplicity, robust classifiers trained with adversarial training achieve state-of-the-art levels of robustness against various newer attacks~\citep{athalye2018obfuscated,croce2020reliable}. For this reason, adversarial training has become one of the most common defenses.

\begin{wrapfigure}{l}{0.5\textwidth}
    \vspace{-0.48cm}
    \caption{Validation accuracy of a model adversarially trained on PGD $L_2$-perturbed CIFAR10 with a ResNet18, evaluated on PGD $L_2$ and Carlini \& Wagner (CW) $L_2$ attacks. Curves are smoothed with exponential moving averaging (weight 0.7).}\label{fig:L2_CW_PGD_tradeoff}
    \includegraphics[width=0.48\textwidth]{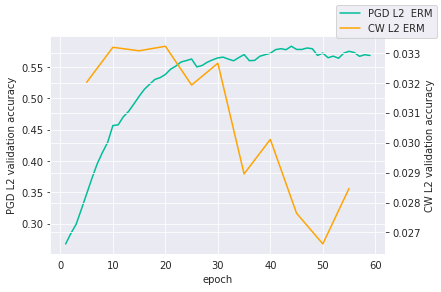}
\end{wrapfigure}

However, \citet{khoury2018geometry} and \citet{tramer2019adversarial} show how training on PGD with a search region constrained by a $p$-norm may not yield robustness against PGD attacks using other $p$-norms. One reason is that different radii are typically chosen for different norms, leading to the search spaces of PGD with respect to different norms to potentially have some mutually exclusive regions. Another reason is that different attacks, such as PGD and the Carlini and Wagner~\citep{carlini2017towards} attacks, optimise different losses (note that this is also true for PGD of different norms). As an example, Fig. \ref{fig:L2_CW_PGD_tradeoff} illustrates how, when training adversarially a model on $L_2$-norm PGD, the accuracy against one attack may improve while it may decrease against another attack, even if the attacks use the same $p$-norm.


Highlighting the need for methods specific to defending against multiple of perturbations, \citet{tramer2019adversarial} select a set of 3 attacks $\mathcal{A} =  \{P_\infty, P_2, P_1\}$, where $P_p$ is PGD with a search region constrained by the $L_p$ norm. They attempt two strategies: the \textbf{average (Avg) strategy} consists in training over all attacks in $\mathcal{A}$ for each input $(x, y)$ in the dataset, and the max strategy, which trains on the attack with the highest loss for each sample:
\begin{align}
    L_{\text{Avg}}(\theta, \mathcal{A}) &= \mathbb{E} \frac{1}{\absval{\mathcal{A}}}\sum_{A\in\mathcal{A}} \ell(\theta, A(x), y) \\
    L_{\text{max}}(\theta, \mathcal{A}) &= \mathbb{E} \max_{A\in\mathcal{A}} \ell(\theta, A(x), y)
\end{align}

\citet{maini2020adversarial} propose a modification to the max method: instead of 3 different PGD adversaries that each iterate over a budget of iterations as in eq. \ref{eq:PGD}, they design an attack consisting in choosing the worst perturbation among $L_\infty$, $L_2$ and $L_1$ PGD every iteration through the chosen number of iterations. This attack, \textbf{Multi-Steepest Descent (MSD)}, differs from the max approach of \citet{tramer2019adversarial} where each attack is individually iterated through the budget of iterations first, and the one leading to the worst loss is chosen at the end. Note that this implies that technically, unlike \citep{tramer2019adversarial}'s Avg approach, \emph{MSD\footnote{In the rest of the paper, we will use MSD to refer to both the MSD attack, and training on MSD as a defense.} only consists in training on a single attack.} \citet{maini2020adversarial} show that, in their experimental setup, MSD yields superior performance to both the Avg and Max approaches.

Nevertheless, there is still a very large gap between the performance of such approaches against data perturbed by ensembles of attacks, and the accuracy on the unperturbed data. 
In order to help address this large gap, we will be exploiting a connection between our goal and domain generalisation.


\subsection{Robustness as a domain generalisation problem}

\textbf{Domain generalisation --}  Out-of-Distribution generalisation (OoD) is an approach to dealing with (typically non-adversarial) distributional shifts.
In the domain generalisation setting, the training data is assumed to come from several different domains, each with a different data distribution.
The goal is to use the variability across training (or seen) domains to learn a model that can generalise to unseen domains while performing well on the seen domains. In other words, the goal is for the model to have consistent performance by learning to be invariant under distributional shifts.
Typically, we also assume access to domain labels, i.e.\ we know which domain each data point belongs to. Many methods for domain generalisation have been proposed -- see~\citep{wang2021generalizing} for a survey.

Our work views adversarial robustness as a domain generalisation problem, where the domains stem from different adversarial attacks. Because different attacks use different methods of searching for adversarial examples, and sometimes different search spaces, they may produce different distributions of adversarial examples\footnote{Another way to think about this, is that if different attacks or tunings yielded identical distributions, then standard results from statistical learning theory would imply similar performance on the various attacks.}. One might draw an analogy to \citet{hendrycks2019robustness}'s work on natural pertubations, where both the type and the strength of the perturbations play a similar role as varying the attacks or their tuning, respectively. There are several reasons why the domains we consider may be distributionally shifted with one another (although the distributions may have some overlap). To non-exhaustively name a few, first, we already evoked how different $p$-norms affect the distributions of adversarial examples yielded by PGD \citep{khoury2018geometry, tramer2019adversarial}. Second, different attacks may optimise different losses -- for example when comparing $P_2$ and $L_2$ CW -- which may yield different solutions. Third, the same attack tuned differently (e.g. different $\epsilon$ or iteration budget) may yield different distributions of adversarial examples since they do not have the same support. Therefore, robustness to attacks unseen during training means robustness against the corresponding distributional shifts at test time. It is natural to frame adversarial robustness as a domain generalisation problem, as we seek a model that is robust to \textit{any} method to generate adversarially distributional shifts within a threat model, including novel attacks.

In spite of this intuition, it is not obvious that such methods would work in the case of adversarial machine learning. First, recent work demonstrates that domain generalisation methods often fail to improve upon the standard \textbf{empirical risk minimisation (ERM)}, i.e.\ minimising loss on the combined training domains without making use of domain labels~\citep{gulrajani2020search}.
On the other hand, success may depend on choosing a method appropriate for the type of shifts at play.
Second, a key difference with most work in domain generalisation, is that when adversarially training, the training distribution shifts every epoch, as the attacks are computed from the continuously-updated values of the weights. In contrast, in domain generalisation, the training domains are usually fixed. Non-stationarity is known to cause generalisation failure in many areas of machine learning, notably reinforcement learning~\citep{igl2020impact}, thereby potentially affecting the success of domain generalisation methods in adversarial machine learning. Third, MSD does not generate multiple domains, which domain generalisation approaches would typically require.

We note that interestingly, the Avg approach of \citet{tramer2019adversarial} can be interpreted as doing domain generalisation with ERM over the 3 PGD adversaries as training domains. Similarly, the max approach consists in applying the Robust Optimisation approach on the same set of domains. Furthermore, \citet{song2018improving} and \citet{bashivan2021adversarial} propose to treat the clean and PGD-perturbed data as training and testing domains from which some samples are accessible during training, and adopt domain adaptation approaches. Therefore, it is difficult to predict in advance how much a domain generalisation approach can successfully improve adversarial defenses.

In this work, we apply the method of \textbf{variance-based risk extrapolation (REx)}~\citep{krueger2021out}, which simply adds as a loss penalty the variance of the ERM loss across different domains.
This encourages worst-case robustness over more extreme versions of the shifts (here, shifts are between different attacks) observed between the training domains.
This can be motivated in the setting of adversarial robustness by the observation that adversaries might shift their distribution of attacks to better exploit vulnerabilities in a model. In that light, REx is particularly appropriate given our objective of mitigating trade-offs in performance between different attacks to achieve a more consistent degree of robustness. We note that our implementation of REx has the same computational complexity per epoch as the MSD, Avg and max approaches, requiring the computation of 3 adversarial perturbations per sample.


\section{Methodology}\label{sec:methodology}
\textbf{Threat model --} In this work, we consider white-box attacks, which are typically the strongest type of attacks as they assume the attacker has access to the model and its parameters. Additionally, the attacks considered in the evaluations are gradient-based, with the exception of AutoAttack, which is composite and includes gradient-free perturbations~\citep{croce2020reliable}. Because we assume that the attacker has access to all of these attacks, we emphasise that, as argued by \citet{athalye2018obfuscated}, the robustness against the ensemble of the different attacks is a better metric for how the defenses perform than the accuracy on each individual attack. Thus, using $\ell_{01}$ as the 0-1 loss, we evaluate the performance on an ensemble of domains $\mathcal{D}$ as:
\begin{align}
    \mathcal{R} = 1 - \mathbb{E} \max_{D\in\mathcal{D}} \ell_{01}(\theta, D(x), y)
\end{align}
\textbf{REx --} We propose to regularise the average loss over a set of training domains $\mathcal{D}$ by the variance of the losses on the different domains:
\begin{align}\label{eq:REx_loss}
    L_{\text{REx}}(\theta, \mathcal{D}) = L_{\text{Avg}}(\theta, \mathcal{D}) 
                                    + \beta \Var_{D\in\mathcal{D}} \mathbb{E}\ \ell(\theta, D(x), y) 
\end{align}
where $\ell$ is the cross-entropy loss. We start penalising by the variance over the training domains once the baseline's accuracies on the seen domains stabilise or peak. 

\textbf{Datasets and architectures --} We consider two datasets: MNIST~\citep{lecun1998gradient} and CIFAR10~\citep{krizhevsky2009learning}. It is still an open problem to obtain high robustness against multiple attacks on MNIST~\citep{tramer2019adversarial, maini2020adversarial}, even at standard tunings of some commonly used attacks. On MNIST, we use a 3-layer perceptron of size [512, 512, 10]. On CIFAR10, we use the ResNet18 architecture~\citep{he2016deep}. We choose two significantly different architectures to illustrate that our approach may work agnostically to the choice of architecture. We always use batch sizes of 128 when training.

\textbf{Optimiser --} We use Stochastic Gradient Descent (SGD) with momentum $0.9$. In subsections \ref{subsec:MNIST} and \ref{apdx:subsec:CIFAR10} we do not perform hyperparameter optimisation, to isolate the effect of REx from interactions with hyperparameter tuning, which would differ for each defense. We use a fixed learning rate of $0.01$ and no weight decay. We fix the coefficient $\beta$ in the REx loss. In subsection \ref{subsec:tuned_CIFAR10}, we optimise hyperparameters. Based on~\citep{rice2020overfitting} and~\citep{pang2020bag}, we use in all cases a weight decay of $5\cdot 10^{-4}$ and a piecewise learning rate decay. For every defense, we search for an optimal epoch to decay the learning rate, with a particular attention to MSD and MSD+REx due to observing a high sensitivity to the choice of learning rate decay milestone. Note that in the case of REx defenses, we always use checkpoints of corresponding baselines before the learning rate is decayed, as we observed this to lead to better performance.

\textbf{Domains --} We consider several domains: unperturbed data, $L_1$, $L_2$ and $L_\infty$ PGD (denoted $P_1, P_2, P_\infty$), $L_2$ Carlini \& Wagner (CW$_2$)~\citep{carlini2017towards}, $L_\infty$ DeepFool (DF$_\infty$)~\citep{moosavi2016deepfool} and AutoAttack (AA) 
 \citep{croce2020reliable}. We use the Advertorch implementation of these attacks~\citep{ding2019advertorch}. For $L_\infty$ PGD, CW and DF, we use two sets of tunings, see appendix \ref{section:apdx:more} for details. The attacks with a $\bullet$ superscript indicate a harder tuning of these attacks that no model was trained on. Those tunings are intentionally chosen to make the attacks stronger. The set of domains \textbf{unseen by all models} is defined as $\{P_\infty^\bullet, DF_\infty^\bullet, CW_2^\bullet, \textit{AutoAttack}_\infty\}$, with additionally \textit{AutoAttack}$_2$ in subsection \ref{subsec:tuned_CIFAR10}. The set of domains \textbf{unseen by a specific model} is the set of all domains except those seen by the model during training, and therefore varies between baselines. We perform 10 attack restarts per sample to reduce randomness in the test set evaluations.

\vspace{-0.1cm}
\textbf{Defenses --} Aside from the adversarial training baselines on PGD of $L_1, L_2$ and $L_\infty$ norms, we define 3 sets of seen domains: $\mathcal{D} = \{\varnothing, P_\infty, DF_\infty, CW_2\}$, $\mathcal{D}_{\textit{PGDs}} = \{\varnothing, P_1, P_2, P_\infty\}$ and $\mathcal{D}_{\textit{MSD}} = \{\textit{MSD}\}$ where $\varnothing$ represents the unperturbed data. We train two Avg baselines: one on $\mathcal{D}$ and one on $\mathcal{D}_{\textit{PGDs}}$. We train the MSD baseline on $\mathcal{D}_{\textit{MSD}}$. We use REx on the Avg baselines on the corresponding set of seen domains. However, when REx is used on the model trained with the MSD baseline, we revert to using the set of seen domains $\mathcal{D}_{\textit{PGDs}}$. While the MSD baseline does not exactly train over $P_1, P_2$ and $P_\infty$ but rather a composition of these three attacks, we use these attacks when applying REx to the MSD baseline as MSD would only generate one domain, which would not allow us to compute a variance over domains. Note that we chose different sets of seen domains, and different baselines (Avg and MSD), in order to show that REx yields benefits on several multi-perturbation baselines, or within a same baseline with different choices of seen domains. We use \textbf{cross-entropy} for all defenses.

\vspace{-0.1cm}
See Appendix \ref{section:apdx:more} for more details about the methodology, such as attack tunings.


\section{Results} \label{sec:results}
In this section, we first illustrate the differences in distributions stemming from different families or tunings of attacks by training an attack classifier in subsection \ref{subsec:attack_discriminator}. We then present our results on MNIST and CIFAR10 in subsections \ref{subsec:MNIST} and \ref{subsec:tuned_CIFAR10}.
We find that REx consistently improves the baselines over ensembles of seen or unseen attacks. REx often sacrifices the performance on the best performing domains (usually, the unperturbed data, $P_1$ and $P_2$) but improves the worst-case accuracy by a much wider margin. This is the case even when using REx on MSD. Also, we find that in general, REx is relatively robust to the choice of $\beta$, albeit it can be used to tune relative performance between easier and harder domains (see Appendix \ref{apdx:sec:varying_beta}). Additionally, more results on the interaction between the advantage in using REx and hyperparameter tuning, the relative performance of REx models on CIFAR10-C and for transfer learning, and more, can be found in Apdx \ref{apdx:sec:additional_results}. More implementation details, observations and results can generally be found in the appendix (see \hyperref[apdx:toc]{Table of Content}).

\begin{figure}[h]
    \vspace{-0.3cm}
    \centering
    \caption{Confusion matrix of a discriminator between attacks (normalised by row).}\label{fig:ViT_discriminate_all_attacks}
    \includegraphics[width=0.8\linewidth]{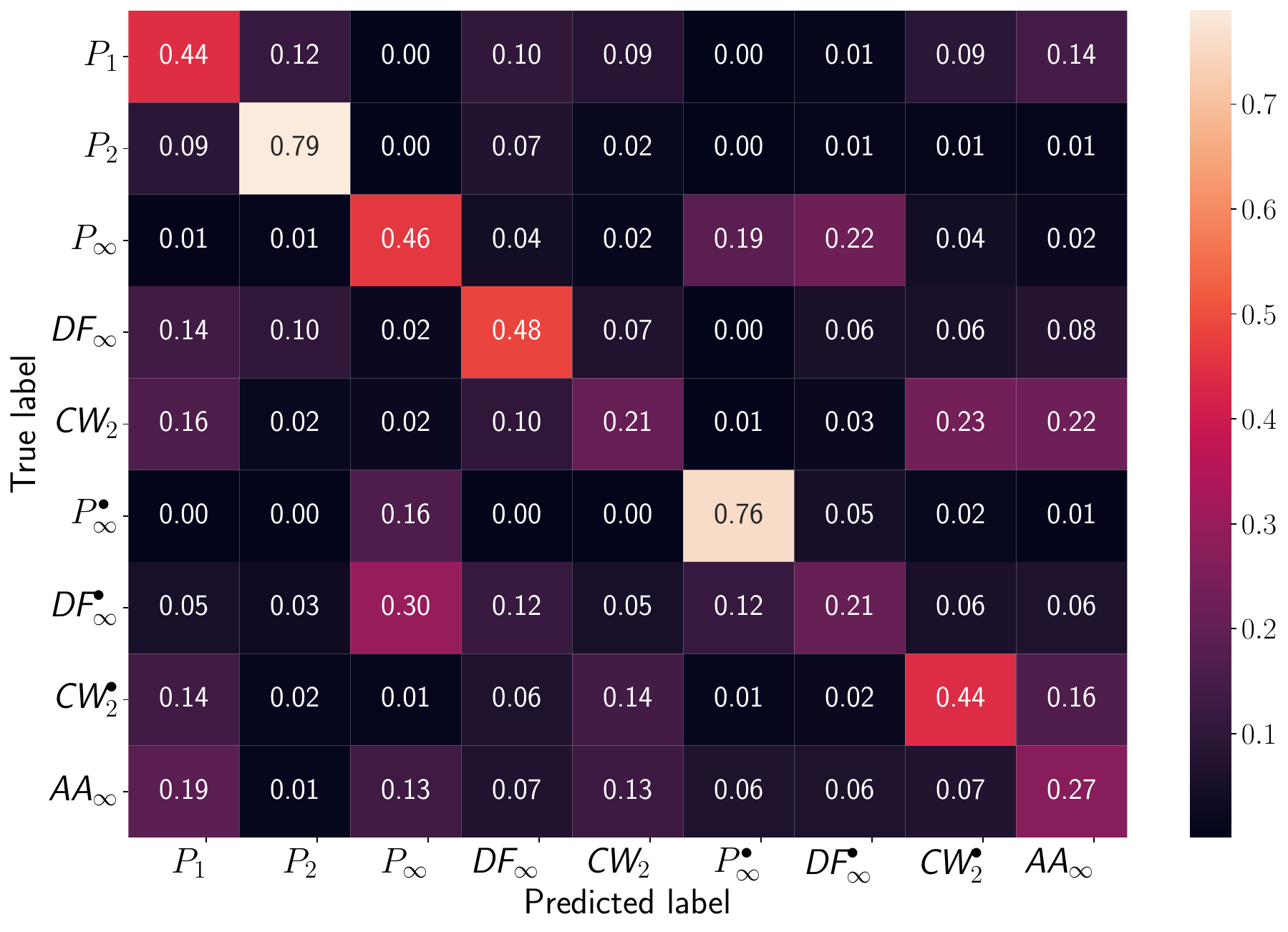}
    \vspace{-0.8cm}
\end{figure}
\subsection{Attacks as different domains}\label{subsec:attack_discriminator}
Before presenting our robustness results, to illustrate how various types or tunings of attacks may correspond to different distributions, we finetune an ImageNet-pretrained vision transformer (ViT) \citep{dosovitskiyimage} to predict which attack was used on the training set of CIFAR10, and test it on the perturbed CIFAR10 test set. Fig. \ref{fig:ViT_discriminate_all_attacks} illustrates how without much engineering effort, the ViT is able to tell apart the distribution of perturbations induced by the various attacks with 45.1\% accuracy, relative to a random chance of $1/9 \simeq 0.11$. To claim unequivocally that all attacks considered are distributionally shifted with respect to one another, we require $p(A_\textit{true}) > p(A\neq A_\textit{true})$ where $A_\textit{true}$ is the true attack used to generate the sample and $p(A)$ is the probability predicted by the model that a perturbation corresponds to attack $A$. This is true for all attacks, except two: $\textit{CW}_2$ and $\textit{DF}_\infty^\bullet$.

The cases of $\textit{CW}_2$ and $\textit{DF}_\infty^\bullet$ (the latter sharing $\epsilon=8/255$ with $P_\infty$ and $\textit{AA}_\infty$) warrant an additional discussion. The former is due to unsuccessful $\textit{CW}_2$ iterations stopping early, i.e. $\Tilde{x} \simeq x$. This problem disappears when such unsuccessful $\textit{CW}_2$ perturbations are no longer involved. This is why the stronger $\textit{CW}_2^\bullet$ is classified more easily. As for $\textit{DF}_\infty^\bullet$ confused with $P_\infty$, we direct the reader to Apdx \ref{apdx:subsec:attack_discriminator} for more details, where we show how this problem vanishes when training a binary $\textit{DF}_\infty^\bullet$ vs $P_\infty$ discriminator.

\begin{tcolorbox}[leftrule=1.5mm,top=0.8mm,bottom=0.5mm]
\textbf{Key observation 1:}
Different types or tunings of attacks induce different distributional shifts that a discriminator can identify to some extent, even for the same choice of $p$-norm and $\epsilon$.
\end{tcolorbox}

\subsection{MNIST}\label{subsec:MNIST}
\begin{table*}[h]
\centering
\caption{Accuracy on MNIST for different domains. Highlighted cells indicate that the domain (row) was used during training by the defense (column). Bold numbers indicate an improvement of at least 1\% accuracy over the baseline used to pretrain REx. Ensembles omit $P_\infty^\bullet$ due to it being overtuned (i.e. tuned to be too strong.).}\label{table:mnist}
\resizebox{0.99\textwidth}{!} {
\begin{tabular}{@{}p{25mm}cccccccccc<{\kern-\tabcolsep}}
\hline
                                   & \multicolumn{10}{c}{Defenses}                                           \\ \hline
                                   & \text{None}                         & \multicolumn{3}{c}{Adversarial training}                                                   & \text{Avg}                          & \text{Avg+REx}                               & $\text{Avg}_{\textit{PGDs}}$                   & $\text{Avg+REx}_{\textit{PGDs}}$                        & \text{MSD}                          & \text{MSD+REx}                               \\ \hline
No attack                              & \cellcolor[HTML]{DDEBF7}98.1 & 98.5                         & 98.3                         & 84.4                         & \cellcolor[HTML]{DDEBF7}99.0 & \cellcolor[HTML]{DDEBF7}90.0          & \cellcolor[HTML]{DDEBF7}98.8 & \cellcolor[HTML]{DDEBF7}87.3          & 88.4                         & \cellcolor[HTML]{DDEBF7}\textbf{90.2} \\
$P_1$                                  & 95.5                         & \cellcolor[HTML]{DDEBF7}96.8 & 96.8                         & 44.0                         & 90.3                         & 72.6                                  & \cellcolor[HTML]{DDEBF7}95.6 & \cellcolor[HTML]{DDEBF7}82.5          & \cellcolor[HTML]{DDEBF7}82.2 & \cellcolor[HTML]{DDEBF7}\textbf{86.8} \\
$P_2$                                  & 1.8                          & 17.7                         & \cellcolor[HTML]{DDEBF7}63.5 & 10.0                         & 53.6                         & 44.0                                  & \cellcolor[HTML]{DDEBF7}68.3 & \cellcolor[HTML]{DDEBF7}\textbf{72.8} & \cellcolor[HTML]{DDEBF7}61.1 & \cellcolor[HTML]{DDEBF7}\textbf{71.8} \\
$P_\infty$                             & 0.0                          & 0.0                          & 2.2                          & \cellcolor[HTML]{DDEBF7}59.2 & \cellcolor[HTML]{DDEBF7}67.7 & \cellcolor[HTML]{DDEBF7}\textbf{70.1} & \cellcolor[HTML]{DDEBF7}58.0 & \cellcolor[HTML]{DDEBF7}\textbf{70.8} & \cellcolor[HTML]{DDEBF7}19.3 & \cellcolor[HTML]{DDEBF7}\textbf{67.4} \\
$DF_\infty$                            & 3.3                          & 5.7                          & 85.9                         & 78.1                         & \cellcolor[HTML]{DDEBF7}92.9 & \cellcolor[HTML]{DDEBF7}84.6          & 92.3                         & 80.9                                  & 56.7                         & \textbf{82.4}                         \\
$CW_2$                                 & 4.4                          & 6.9                          & 56.5                         & 62.3                         & \cellcolor[HTML]{DDEBF7}68.8 & \cellcolor[HTML]{DDEBF7}68.3          & 59.9                         & 41.4                                  & 77.1                         & 47.3                                  \\ 
$DF_\infty^\bullet$                            & 0.0                          & 0.0                          & 0.0                          & 19.4                         & 7.1                          & \textbf{64.8}                         & 3.7                          & \textbf{58.4}                         & 15.8                         & \textbf{19.9}                         \\
$CW_2^\bullet$                                 & 2.3                          & 2.8                          & 16.0                         & 30.2                         & 23.2                         & \textbf{42.1}                         & 16.4                         & 12.1                                  & 40.2                         & 12.9                                  \\
AutoAttack$_\infty$                        & 0.0                          & 0.0                          & 0.1                          & 55.0                         & 42.3                         & \textbf{58.8}                         & 34.9                         & \textbf{40.6}                         & 1.5                          & \textbf{31.2}                         \\ \hline
Ensemble (seen)                        & -                            & -                            & -                            & -                        & 63.2                         & 63.4                                  & 55.5                         & \textbf{64.5}                         & 19.3                         & \textbf{60.1}                         \\
Ensemble (unseen by all models) & 0.0                          & 0.0                          & 0.0                          & 9.3                         & 3.4                          & \textbf{34.6}                         & 1.2                          & \textbf{8.1}                          & 0.6                          & \textbf{3.9}                          \\
Ensemble (unseen by this model)    & 0.0                          & 0.0                          & 0.0                          & 2.7                         & 3.4                          & \textbf{25.9}                         & 1.2                          & \textbf{8.1}                          & 0.6                          & \textbf{3.9}                          \\
Ensemble (all)           & 0.0                          & 0.0                          & 0.0                          & 2.7                         & 3.4                          & \textbf{25.9}                         & 1.2                          & \textbf{8.1}                          & 0.6                          & \textbf{3.9}                          \\ \hline
$P_\infty^\bullet$                             & 0.0                          & 0.0                          & 0.0                          & 5.1                         & 0.6                          & \textbf{4.0}                                   & 0.9                          & 0.7                                   & 0.2                          & 1.0                                   \\ \hline
\end{tabular}
}
\end{table*}
We report our multiperturbation robustness results on MNIST in Table \ref{table:mnist}. REx significantly improves robustness against the ensembles of attacks, whether seen or unseen, and in particular on $P_\infty$ and AutoAttack. REx also yields notable improvements against all ensembles, seen or unseen, when used on the Avg baselines. Note however that as in domain generalisation, when used on all baselines except MSD, REx sacrifices performance on the best performing seen domains in order to improve the performance on the strongest attacks. We believe that this trade-off may be worth it for applications where robustness is critical, as for example the 9\% of clean accuracy lost by using REx on one Avg baseline translates in an increase of robustness from 3.4\% to 25.9\% on the ensemble of all attacks excluding the overtuned (i.e. tuned too strongly, leading to negligible accuracy) $P_\infty^\bullet$ adversary.

Our test with tuning the $P_\infty^\bullet$ adversary with $\epsilon = 0.4$ instead of the common tuning of 0.3 on MNIST suggests that REx does not rely on gradient masking\citep{athalye2018obfuscated} compared to the baselines, as the accuracy drops to near 0 values for all models, showing that attacks are successfully computed. This is reinforced by the REx models' AutoAttack performance. A second observation is that the MSD baseline performs surprisingly poorly against AutoAttack and $P_\infty$. We note that experiments with a learning rate schedule (not reported here) did not significantly improve performance of MSD, ruling out the absence of schedule as a cause. While we use the original code of \citet{maini2020adversarial}, this could be because we did not use the same architecture as them on MNIST. Furthermore, \citet{maini2020adversarial} did not evaluate on AutoAttack as their work predates the publication of \citet{croce2020reliable}. In any case, the MSD model did not achieve substantial robustness against $P_\infty$ and AutoAttack in our experiments. This leads to poor performance against all ensembles of attacks, whether seen or unseen, as those include either $P_\infty$ or AutoAttack adversaries. Finally, a third observation is that $P_\infty$ training performs remarkably well in this experiment on the ensemble of attacks, compared to the multi-perturbation baselines.

\begin{tcolorbox}[leftrule=1.5mm,top=0.8mm,bottom=0.5mm]
\textbf{Key observation 2 (MLP on MNIST):}
REx improves performance of all baselines on MNIST with a multilayer perceptron, from 3.4\% with the best baseline to 25.9\% accuracy against an ensemble of $L_p$ attacks, sacrificing a little robustness against the weakest attacks.
\end{tcolorbox}

\begin{figure*}[h]
  \begin{subfigure}[t]{.49\textwidth}
    \centering
    \includegraphics[width=\linewidth]{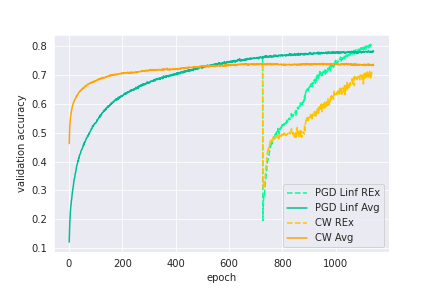}
    \caption{MNIST seen attacks}
  \end{subfigure}
  \hfill
  \begin{subfigure}[t]{.49\textwidth}
    \centering
    \includegraphics[width=\linewidth]{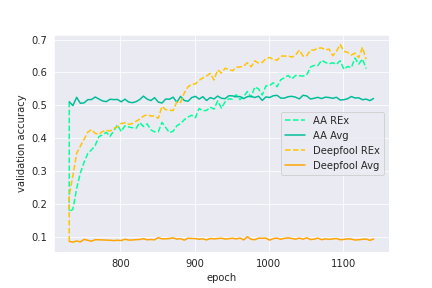}
    \caption{MNIST unseen attacks}
  \end{subfigure}

  \caption{Validation accuracy of Avg on MNIST with and without REx (dashed line), against seen attacks (left) and unseen attacks (right) (AA=AutoAttack).}\label{fig:validation_with_REx_MNIST}
  \vspace{-0.5cm}
\end{figure*}
\subsection{Early stopping and REx}
In Figure \ref{fig:validation_with_REx_MNIST}, we observe how regularising by the variance over the domains leads to better peak performance when using REx, over different baselines. Like the baselines, we early stop REx models. For all defenses, we use the validation set to choose when to early stop, by selecting the epoch when the performance peaks on the ensemble of seen domains. As shown on the curves, even though we stop training before REx reaches higher performance on the seen attacks, we still get significant improvements on the individual unseen attacks and against the ensembles, as reported in Table \ref{table:mnist}. See Appendix \ref{section:apdx:more} for more details on how REx is used, and Appendices \ref{apdx:subsec:CIFAR10} and \ref{apdx:sec:when_to_early_stop} for more details on how, in contrast with MNIST with the MLP, early stopping is required much earlier and the validation curves no longer behave monotonically on a different dataset and architecture.

\subsection{CIFAR10 with hyperparameter optimisation}\label{subsec:tuned_CIFAR10}
On CIFAR10 with the ResNet18, we use weight decay and search for an optimal learning rate schedule individually for each defense. The results are summarised in Table \ref{table:tuned_cifar}. We chose not to use the Avg$_\textit{PGDs}$ baseline here, due to performing worse compared to the other baselines on CIFAR10 with a ResNet18 (which we confirm to be especially true when tuning hyperparameters in preliminary experiments). We direct the reader to Apdx \ref{apdx:subsec:CIFAR10} for CIFAR10 results with an ablation on custom hyperparameter optimisation for each model, which leads to even greater advantages in using REx, and which includes the Avg$_\textit{PGDs}$ baseline. 

\begin{wraptable}{r}{8.3cm}
\vspace{-0.33cm}
\caption{Accuracy on CIFAR10, with hyperparameter tuning. Ensembles omit $\textit{CW}_2^\bullet$ due to overtuning.}\label{table:tuned_cifar}
\resizebox{0.6\textwidth}{!} {
\begin{tabular}{@{}p{25mm}ccccc<{\kern-\tabcolsep}}
\hline
                                   & \multicolumn{5}{c}{Defenses}                                           \\ \hline
                                   & \text{$P_\infty$}                                                   & \text{Avg}                          & \text{Avg+REx}                                  & \text{MSD}                          & \text{MSD+REx}                               \\ \hline
No attack                           & 80.8                         & \cellcolor[HTML]{DDEBF7}80.0 & \cellcolor[HTML]{DDEBF7}76.8          & 78.6 & \cellcolor[HTML]{DDEBF7}77.4          \\
$P_1$                               & 78.2                         & 78.3                         & 74.9                                  & \cellcolor[HTML]{DDEBF7}76.6 & \cellcolor[HTML]{DDEBF7}75.2          \\
$P_2$                               & 70.0                         & 67.9                         & 68.7                                  & \cellcolor[HTML]{DDEBF7}69.8 & \cellcolor[HTML]{DDEBF7}68.7          \\
$P_\infty$                          & \cellcolor[HTML]{DDEBF7}47.3 & \cellcolor[HTML]{DDEBF7}34.4 & \cellcolor[HTML]{DDEBF7}\textbf{48.1} & \cellcolor[HTML]{DDEBF7}45.8 & \cellcolor[HTML]{DDEBF7}\textbf{48.3} \\
$DF_\infty$                         & 69.0                         & \cellcolor[HTML]{DDEBF7}64.4 & \cellcolor[HTML]{DDEBF7}\textbf{67.1} & 67.1                         & 67.3                                  \\
$CW_2$                              & 17.4                         & \cellcolor[HTML]{DDEBF7}14.5  & \cellcolor[HTML]{DDEBF7}\textbf{29.6} & 17.9                         & \textbf{20.9}                         \\
$P_\infty^\bullet$                  & 28.9                         & 16.9                         & \textbf{28.2}                         & 27.4                         & \textbf{30.7}                         \\
$DF_\infty^\bullet$                 & 46.3                         & 35.2                         & \textbf{45.3}                         & 44.9                         & \textbf{46.2}                         \\
AutoAttack$_\infty$                 & 44.8                         & 33.5                         & \textbf{43.1}                         & 42.8                         & \textbf{44.8}                         \\
AutoAttack$_2$                      & 57.7                         & 59.2                         & 58.4                                  & 61.1                         & 56.6                                  \\
\hline
Ensemble (seen)                     & -                            & 14.5                          & \textbf{29.2}                         & 45.8                         & \textbf{48.2}                         \\
Ensemble (unseen by all models)     & 28.9                         & 16.9                         & \textbf{27.9}                         & 27.4                         & \textbf{30.3}                         \\
Ensemble (unseen by this model)     & 16.9                         & 16.9                         & \textbf{27.9}                         & 16.5                         & \textbf{19.6}                         \\
Ensemble (all)                      & 16.9                         & 14.2                          & \textbf{23.5}                         & 16.5                         & \textbf{19.6}                         \\ \hline
$CW_2^\bullet$                      & 2.5                          & 4.8                          & 5.3                                   & 1.9                          & \textbf{3.7}                          \\ \hline
\end{tabular}
}
\vspace{-0.3cm}
\end{wraptable}

The cumulative effect of weight decay and learning rate schedules significantly improves all defenses' robustness, as shown by \citet{rice2020overfitting} and \citet{pang2020bag}. Once again, we observe that REx improves significantly the seen and unseen ensemble accuracies over the baselines. 
We also note that $P_\infty$ adversarial training performs better than the baselines on the ensemble of attacks used in this paper, even with the addition of AutoAttack$_2$ to the ensembles containing unseen attacks. Moreover, only REx is able to perform better than $P_\infty$ adversarial training on $P_\infty$ attacks. In other words, multi-perturbation defenses only perform better than $P_\infty$ against ensembles of attacks when used with REx.

While MSD performs significantly better with a ResNet18 on CIFAR10 than with the MLP on MNIST (likely due to using the same architecture as them on CIFAR10), as suspected when discussing our optimiser methodology in Sec. \ref{sec:methodology}, there are interaction effects between hyperparameter tuning and the performance of REx relative to a baseline. Improvements of MSD+REx over MSD are sensitive to hyperparameter tuning, specifically at which epoch to start using REx, and when to decay the learning rate. This sensitivity, and the lower advantage of MSD+REx over MSD, is likely due to the fact that the MSD baseline does not train over multiple domains. REx was originally designed to be used with a baseline using ERM on multiple domains as loss function. Therefore, when it is used in tandem with MSD, REx uses the loss indicated in eq. \ref{eq:REx_loss}. However, because as mentioned before, the Avg$_\textit{PGDs}$ baseline performs significantly worse than the MSD baseline, it is likely that the advantage in using REx is impacted negatively by the suboptimality of the first (ERM) term in the REx loss. Nevertheless, the variance penalty is beneficial enough to achieve higher robustness with MSD+REx than MSD.

\begin{wraptable}{l}{8.5cm}
\caption{Accuracy on two non-$L_p$ perturbations to CIFAR10.}\label{table:tuned_cifar_non_Lp}
\resizebox{0.6\textwidth}{!} {
\begin{tabular}{@{}p{18mm}ccccc<{\kern-\tabcolsep}}
\hline
                                   & \multicolumn{5}{c}{Defenses}                                           \\ \hline
                   & $P_\infty$ & Avg & Avg+REx & MSD & MSD+REx \\
                   \hline
RecolorAdv & 50.5                     & 24.5                    & \textbf{63.5}           & 56.0                   & \textbf{58.2}          \\
StAdv      & 12.1                     & 4.0                     & \textbf{31.8}           & 17.6                   & \textbf{22.7}         \\ \hline
\end{tabular}
}
\end{wraptable}
While our results have focused on $L_p$ attacks, we evaluate the tuned CIFAR10 models on two additional non-$L_p$ attacks: RecolorAdv \citep{laidlaw2019functional}, and Spatial Transformations \citep{xiaospatially}. We observe in Table \ref{table:tuned_cifar_non_Lp} that REx provides significantly better robustness against these perturbations than any other baseline.

\begin{wraptable}{r}{8.5cm}
\vspace{-0.4cm}
\centering
\caption[width=0.5\textwidth]{Average accuracy of tuned CIFAR10 models on CIFAR10-C corruptions.}\label{table:averages_cifar-c}
\resizebox{0.6\textwidth}{!} {
\begin{tabular}{@{}p{12mm}cccccc<{\kern-\tabcolsep}}
\hline
                                   & \multicolumn{6}{c}{Defenses}                                           \\ \hline
                   & None & $P_\infty$ & Avg & Avg+REx & MSD & MSD+REx \\
                   \hline
Average            & 21.8                     & 52.3                     & 26.5                    & 48.2                    & 42.1                   & 51.2        \\ \hline          
\end{tabular}
}
\end{wraptable}
Additionally, we also report results on CIFAR10-C \citep{hendrycks2019robustness} in Table \ref{table:averages_cifar-c}. The dataset consists in mimicking several natural corruptions on CIFAR10 images at various strength, which as argued before, can be seen as a non-adversarial analogue of trying both different types, and tunings of adversarial attacks. While this is not an adversarial robustness benchmark, it shows that REx significantly improves robustness of multiperturbation defenses to non-adversarial shifts it is used on, in spite of what REx models' lower in-distribution clean accuracy on CIFAR10 may have suggested in Table \ref{table:tuned_cifar}. 

We invite readers interested in more results or details to consult Apdx \ref{apdx:sec:additional_results}.

\begin{tcolorbox}[leftrule=1.5mm,top=0.8mm,bottom=0.5mm]
\textbf{Key observations 3 (ResNet18 on CIFAR10):}
\begin{itemize}
    \item REx improves the performance of all baselines on CIFAR10 with a ResNet18, from 16.9\% with the best baseline to 23.5\% accuracy against an ensemble of $L_p$ attacks, by sacrificing a little robustness against the weakest individual attacks.
    \item Multi-perturbation defenses only achieve higher $P_\infty$ and worst-case performance than $P_\infty$ adversarial training when they are used in conjunction with REx.
    \item REx also provides better robustness on some common non-$L_p$ attacks, and significantly improves robustness of baselines it is used on against CIFAR-C's non-adversarial shifts.
\end{itemize}
\end{tcolorbox}

\section{Conclusion}

An attacker seeking to exploit a machine learning model is liable to use the most successful attack(s) available to them.
Thus, defenses against adversarial examples should ideally provide robustness against any reasonable attack, including novel attacks.
In particular, the worst-case robustness against the set of available attacks is most reflective of the performance achieved against a dedicated and sophisticated adversary.

We achieve state-of-the-art worst-case robustness by applying the domain generalisation technique of V-REx~\citep{krueger2021out}, which seeks to equalise performance across attacks used at training time.
Our approach is simple, practical, and effective.
It produces consistent performance improvements over baselines across different datasets, architectures, training attacks, test attack types and tunings. One limitation, as often in adversarial machine learning, is that our results make no \emph{guarantees} about attacks that were not used in the evaluation. Another limitation lies in the slight loss of accuracy on the unperturbed data, albeit we believe the improvements in adversarial and non-adversarial robustness and promising research directions are significant enough to be of interest to the community.

Indeed, our work demonstrates the potential of applying domain generalisation approaches to adversarial robustness.
Future work could investigate other domain generalisation methods, such as Distributionally Robust Optimisation (DRO)~\citep{sagawa2019distributionally} or Invariant Risk Minimisation (IRM)~\citep{arjovsky2019invariant}, and evaluate their effectiveness against more and newer attacks.

Our results are particularly interesting in light of REx's potential failure to improve performance over well-tuned baselines on non-adversarial domain generalisation benchmarks \citep{gulrajani2020search}.

\begin{ack}
Ioannis Mitliagkas acknowledges support by an NSERC Discovery grant (RGPIN-2019-06512), a Samsung grant and a Canada CIFAR AI chair. Irina Rish acknowledges the support from Canada CIFAR AI Chair Program and from the Canada Excellence Research Chairs Program. Pouya Bashivan is supported by Healthy-Brains-Healthy-Lives start-up supplement grant, William Dawson Scholar Award, and NSERC grants DGECR-2021-00300 and RGPIN-2021-03035.

We thank Shoaib Ahmed Siddiqui, Jonathan Uesato, Cassidy Laidlaw, Pratyush Maini and Sven Gowal for enriching discussions that helped us refine the paper. We also thank reviewers of ICLR 2023 and ICML 2023, along with attendees of the New Frontiers in Adversarial Machine Learning workshop for their valuable feedback.
\end{ack}

\newpage

\bibliographystyle{template/icml2023}
\bibliography{biblio}


\newpage
\appendix
\tableofcontents\label{apdx:toc}
\onecolumn
The code can be found at \href{https://github.com/AIproj/Towards-Out-of-Distribution-Adversarial-Robustness}{https://github.com/AIproj/Towards-Out-of-Distribution-Adversarial-Robustness}. \\
\section{More on methodology} \label{section:apdx:more}
\subsection{Motivation} \label{apdx:subsec:motivation}
\textbf{Benchmark design}: we use a similar benchmark as \citep{tramer2019adversarial} and \citep{maini2020adversarial}, with the addition of AutoAttack. Furthermore, to empirically support our claims that REx improves worst case robustness, we additionally consider ensembles of seen, unseen, and all attacks. We consider worst-case ensembles because for real-world robustness, an attacker may try several attacks until they succeed. The evaluation against an ensemble of seen attacks is to support our first claim, i.e. that REx improves multi-perturbation robustness, and the consideration of unseen attacks supports the second claim: that REx generalises to attacks that were not seen during training.

\textbf{Why weak attacks}: weak attacks capture a signal that would be missed otherwise. Indeed, as argued in the discussion of our results, REx does improve significantly robustness against strong attacks and even unbounded attacks (which can't be ignored in the general context of robustness until out-of-distribution detectors are perfectly able to capture shifts beyond the typical balls); however, REx often lowers the accuracy slightly on the weakest attacks. In other words, weak attacks allow us to capture a trade-off in using REx, which is important information for readers. Not having those evaluations would suggest that REx is an improvement in every setting.

\textbf{Unbounded attacks}: most work in the literature focuses on bounded attacks. This is because bounded attacks constrain the strength of attacks, and have long been needed to obtain non-trivial robustness results. In this work, we consider several of the most commonly used bounded attacks, and show the improvement yielded by REx on the ones affecting the accuracy of the model the most. However, at deployment, an attacker might use some unbounded attacks, especially if they are not perceptible. We encourage the reader to decide from Fig. \ref{fig:adv_examples_Avg} and \ref{fig:adv_examples_Avg_REx} whether they consider the corruption to be perceptible by a human, in spite of being out of the $\epsilon = 0.5 \ L_2$ ball. In the absence of algorithms able to detect perturbations larger than the usual choices of $\epsilon$, we argue that non-visually perceptible adversarial attacks, regardless of being bounded or not, are of particular concern. Therefore, we additionally consider an unbounded attack (CW) to show that REx provides benefits even in that less-studied case. Note that we find that REx also improves robustness when bounding CW by rejecting examples out of the $L_2$ ball of radius $\epsilon=0.5$ on CIFAR10.

\textbf{Keeping ``overtuned'' attacks}: in our tables of results, we refer to some attacks are ``overtuned''. What is meant is that after choosing an alternative tuning of those attacks, all models mostly failed against them. For example, for $P^\bullet_\infty$ on MNIST, this is done by choosing an $\epsilon = 0.4$ instead of the commonly used value of $0.3$. Since we already provide a significant number of evaluations throughout this work, we decided to report those overtuned attacks to highlight two points. First, the models still have weakness, and as seen in Fig. \ref{fig:adv_examples_Avg} and \ref{fig:adv_examples_Avg_REx}, adversarial examples produced by the overtuned attacks are not necessarily perceptible, so a truly robust model should defend against them. Second, as argued in the discussion, these attacks, along with the AutoAttack evaluation, highlight that REx does not rely on gradient obfuscation \citep{athalye2018obfuscated}.

\textbf{Ablation on hyperparameter optimisation}: we perform an ablation on hyperparameter optimisation on CIFAR10 in Sec. \ref{apdx:subsec:CIFAR10}. This is done in order to highlight several points. First, that without hyperparameter optimisation, REx-based defenses may outperform hyperparameter-optimised baselines. Second, as argued in the discussion of Sec. \ref{subsec:tuned_CIFAR10}, in the specific case of the MSD baseline, while the use of REx on that baseline still yields improvements over the baseline, it is slightly less pronounced than without hyperparameter optimisation. 

\textbf{Max baseline}: we choose not to include the Max baseline from \citep{tramer2019adversarial} since Tables 3 and 4 of \citep{maini2020adversarial} have evaluations on a very similar benchmark. Note that \citet{maini2020adversarial} find on CIFAR10 that Avg performs significantly better than Max. Moreover, on both MNIST and CIFAR10, \citet{maini2020adversarial} find that MSD performs better than both Avg and Max anyway, so improving on MSD with REx implies that MSD+REx would outperform Max on both datasets.

\subsection{Attack tunings}
Using Advertorch's and Croce's implementation of AutoAttack's\footnote{https://github.com/fra31/auto-attack} parameter names, we report the attacks' tuning here.
\\

\subsubsection{MNIST (subsection \ref{subsec:MNIST})}
\begin{enumerate}
    \item $P_1$: $\epsilon = 10$, $n_{\textit{iter}} = 40$, $\epsilon_\textit{iter} = 0.5$
    \item $P_2$: $\epsilon = 2$, $n_{\textit{iter}} = 40$, $\epsilon_\textit{iter} = 0.1$
    \item $P_\infty$: $\epsilon = 0.3$, $n_{\textit{iter}} = 40$, $\epsilon_\textit{iter} = 0.01$
    \item $\textit{DF}_\infty$: $\epsilon = 0.11$, $n_\textit{iter} = 30$
    \item $\textit{CW}_2$: $\textit{max\_iterations} = 20$, $\textit{learning\_rate} = 0.1$, $\textit{binary\_search\_steps} = 5$
    \item $P_\infty^\bullet$: $\epsilon = 0.4$, $n_{\textit{iter}} = 40$, $\epsilon_\textit{iter} = 0.033$
    \item $\textit{DF}_\infty^\bullet$: $\epsilon = 0.4$, $n_\textit{iter} = 50$
    \item $\textit{CW}_2^\bullet$: $\textit{max\_iterations} = 30$, $\textit{learning\_rate} = 0.12$, $\textit{binary\_search\_steps} = 7$
    \item \textit{AutoAttack}$_\infty$: $\epsilon = 0.3$, $\textit{norm} = ``\text{Linf}"$
\end{enumerate}
The MSD attack uses the same tuning as the individual $P_p$ attacks.

\subsubsection{CIFAR10 without hyperparameter optimisation (subsection \ref{apdx:subsec:CIFAR10})}
\begin{enumerate}
    \item $P_1$: $\epsilon = 10$, $n_{\textit{iter}} = 40$, $\epsilon_\textit{iter} = \frac{2}{255}$
    \item $P_2$: $\epsilon = 0.5$, $n_{\textit{iter}} = 40$, $\epsilon_\textit{iter} = \frac{2}{255}$
    \item $P_\infty$: $\epsilon = \frac{8}{255}$, $n_{\textit{iter}} = 40$, $\epsilon_\textit{iter} = \frac{2}{255}$
    \item $\textit{DF}_\infty$: $\epsilon = 0.011$, $n_\textit{iter} = 30$
    \item $\textit{CW}_2$: $\textit{max\_iterations} = 20$, $\textit{learning\_rate} = 0.01$, $\textit{binary\_search\_steps} = 5$
    \item $P_\infty^\bullet$: $\epsilon = \frac{12}{255}$, $n_{\textit{iter}} = 70$, $\epsilon_\textit{iter} = \frac{2}{255}$
    \item $\textit{DF}_\infty^\bullet$: $\epsilon = \frac{8}{255}$, $n_\textit{iter} = 50$
    \item $\textit{CW}_2^\bullet$: $\textit{max\_iterations} = 30$, $\textit{learning\_rate} = 0.012$, $\textit{binary\_search\_steps} = 7$
    \item \textit{AutoAttack}$_\infty$: $\epsilon = \frac{8}{255}$, $\textit{norm} = ``\text{Linf}"$
\end{enumerate}
MSD uses the same tuning as the individual $P_p$ attacks.

\subsubsection{CIFAR10 with hyperparameter optimisation (subsection \ref{subsec:tuned_CIFAR10})}\label{apdx:subsec:tuned_cifar10_tunings}
We use the same tuning as above for testing, with the addition of an $\textit{AutoAttack}_2$ adversary with $\epsilon = 0.5$ and $\textit{norm}=$``L2''. However, for training, based on \citep{rice2020overfitting}, we set
\begin{enumerate}
    \item $P_1$: $\epsilon = 10$, $n_{\textit{iter}} = 10$, $\epsilon_\textit{iter} = \frac{20}{255}$
    \item $P_2$: $\epsilon = 0.5$, $n_{\textit{iter}} = 10$, $\epsilon_\textit{iter} = \frac{15}{255}$
    \item $P_\infty$: $\epsilon = \frac{8}{255}$, $n_{\textit{iter}} = 10$, $\epsilon_\textit{iter} = \frac{2}{255}$
\end{enumerate}
and do the same for MSD.

\subsection{More on how the attack discriminator is trained}\label{apdx:subsec:methodology_ViT_attack_discriminator}
In order to train the attack discriminator of subsections \ref{subsec:attack_discriminator} and \ref{apdx:subsec:attack_discriminator}, we load the ``IMAGENET1K\_SWAG\_LINEAR\_V1'' weights \citep{singh2022revisiting} of a ViT b16 available on Torchvision with
\begin{verbatim}
torchvision.models.vit_b_16(weights="IMAGENET1K_SWAG_LINEAR_V1") 
\end{verbatim} 
We freeze all but the last (linear) layer of the ImageNet-pretrained ViT, which we reset and adapt to the proper number of output classes, and which is therefore the only layer that we train. We use a ViT because it has significantly higher performance than a ResNet18, which we tried initially and found to have much difficulty converging to good classifiers both on the validation and test sets, even after 150 epochs. In contrast, the ViT almost reaches its final values in 2 epochs.

We proceed by using the attacks with their tuning from subsubsection \ref{apdx:subsec:tuned_cifar10_tunings} on CIFAR10, using the hyperparameter-optimised model adversarially trained against $P_\infty$. Since the task is to show that a discriminator can classify different attacks, we do not need to perturb all CIFAR10 classes and limit ourselves to a single at a time. The confusion matrices provided in this work are based on perturbing all CIFAR10 samples of class "0" (airplanes). After saving all the images of airplanes perturbed with every attack, we load them as a dataset for attack classification with the pretrained ViT. Our code allows one generate or use this dataset, selecting the classes and attacks used. 

As preprocessing, we transform the images into perturbations by subtracting the unperturbed original image from the adversarial images, i.e. $A(x^0) - x^0$ where $x^0$ is the original unperturbed image and $A$ is an attack. Furthermore, we opt to standardise at an instance level (i.e. per image) instead of using batch or dataset statistics. A seed of 0 is used in every case.

We use Adam \citep{kingma2014adam} as an optimiser with its default Pytorch settings in torch1.13\footnote{Every other experiment uses torch1.8.1, the ViT analysis being a late addition to this work and ViT requiring more recent Pytorch versions.} (lr=0.001, betas=(0.9, 0.999), eps=1e-08, weight\_decay=0).

\begin{figure}[hb]
    \centering
    \includegraphics[width=0.76\linewidth]{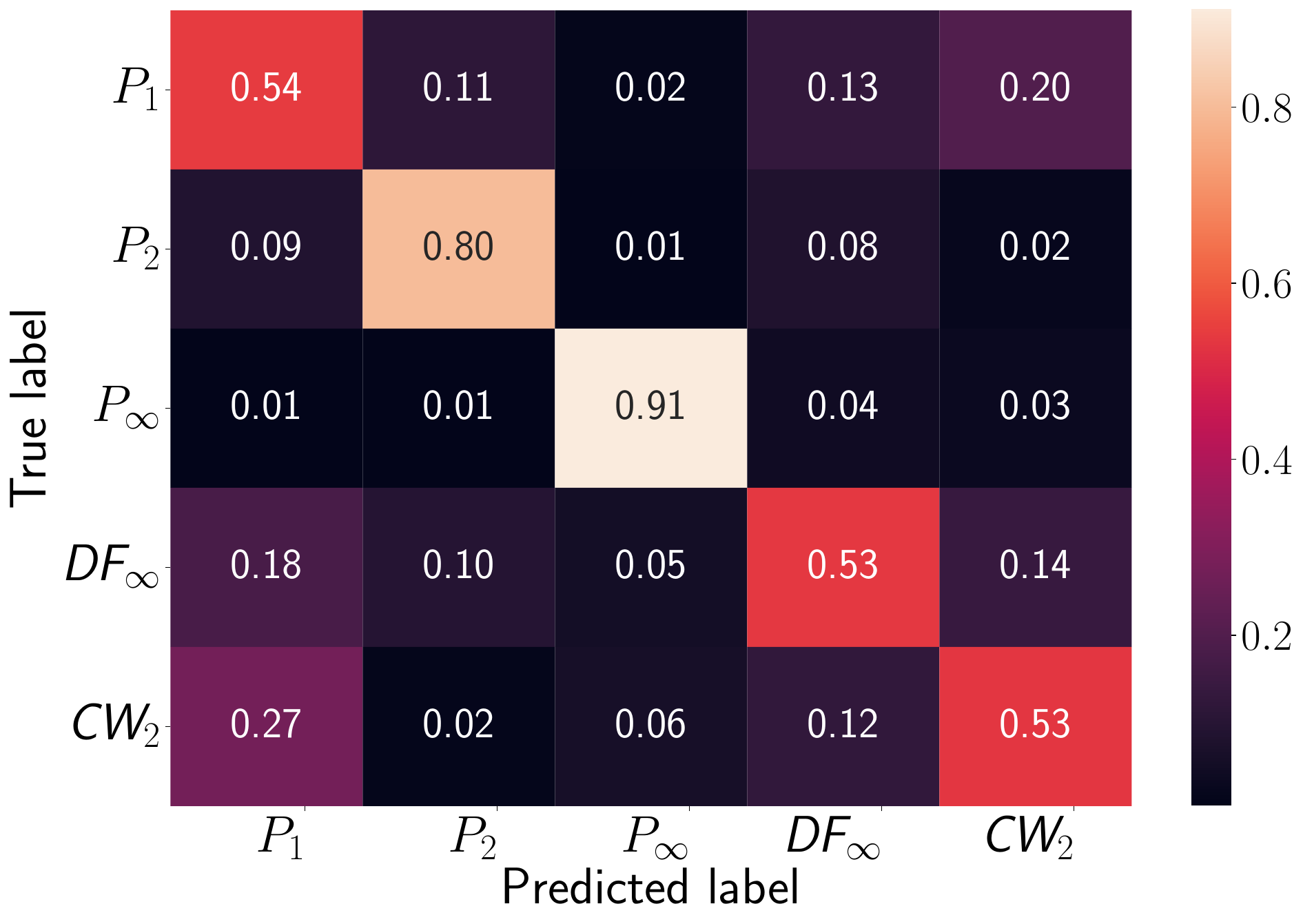}
    \caption{Confusion matrix of ViT predicting whether perturbations stem from 5 different types of attacks (normalised by row). Test accuracy is 66.3\%.}
    \label{fig:ViT_all_no_tunings_no_AA}
\end{figure}

We also report the epoch at which we early stopped training for the analyses:
\begin{enumerate}
    \item Fig. \ref{fig:ViT_discriminate_all_attacks} (discriminating between all $L_p$ attacks considered in this work): epoch 40.
    \item Fig. \ref{fig:ViT_Pinf_DFinfmod} (discriminating between $P_\infty$ and $\textit{DF}_\infty^\bullet$): epoch 16.
    \item Fig. \ref{fig:ViT_all_no_tunings_no_AA} (discriminating between $P_1, P_2, P_\infty, \textit{DF}_\infty, \textit{CW}_2$): epoch 56.
\end{enumerate}
As a reminder, the same base dataset is used for all 3 models, retaining only the samples corresponding to attacks selected for the analysis.

\textbf{Disclaimer:} we made no particular attempt to improve further the attack discriminator. It is probable that more engineering, e.g. transfer learning or preprocessing heuristics, might lead to even better results. However, we believe our current models achieve the intended goal of showing that all these attacks induce different distributions even for the same $p$-norm and $epsilon$, as evidenced by the fact that for any attack, we can show on the test set that $p(A_\textit{true}) > p(A\neq A_\textit{true})$ where $A_\textit{true}$ is the true attack used to generate the sample and $p(A)$ is the probability predicted by the model that a perturbation corresponds to attack $A$. See \ref{subsec:attack_discriminator} and \ref{apdx:subsec:attack_discriminator}.

\subsection{More on how the models are trained}
Note that when activating REx, we always reset the optimiser to avoid using accumulated momentum from the baseline. When tuning hyperparameters, we find that activating the REx penalty after learning rate decays generally is a worse strategy than activating it before when the baseline's accuracy decreases after a decay. All models are trained using a single NVIDIA A100 for MNIST and 2 NVIDIA A100s for CIFAR10.

\subsubsection{No hyperparameter optimisation}
First, we pretrain the architecture on the clean dataset. Then, the baseline is trained on the appropriate seen domains. On MNIST, convergence does not happen in many baselines' case until thousands of epochs. Therefore, we choose to stop training when progress on the seen domains slows, as in Fig. \ref{fig:validation_with_REx_MNIST} where we stopped training for example at epoch 1125 for both the Avg and the Avg+REx models. For CIFAR10, as Fig. \ref{fig:validation_with_REx_CIFAR} indicate, the accuracies peak in significantly less epochs, so we early stop when the Ensemble (seen) accuracy on the seen domains peaks. Note that we do this manually by looking at the seen domains' validation curves (see Sec. \ref{apdx:sec:when_to_early_stop} for more details on early stopping). REx is triggered on a baseline before the baseline's early stopping epoch, when progress on the seen domains slows.

\emph{An important precision about Fig. \ref{fig:validation_with_REx_MNIST}, \ref{fig:validation_with_REx_CIFAR} is that unseen attack performance is only evaluated every 5 epochs, hence the jagged aspect of the curves. We do this because of the huge computational cost of running all 9 attacks on each sample every epoch.}

For a full description of early stopping and when we activated the REx penalty on baselines:

\textbf{MNIST}
\begin{itemize}
    \item $P_1$ model: early stopped at epoch 95
    \item $P_2$ model: early stopped at epoch 75
    \item $P_\infty$ model: early stopped at epoch 1125
    \item Avg model: early stopped at epoch 1125
    \item Avg+REx model: REx penalty activated at epoch 726, early stopped at epoch 1125
    \item Avg$_\textit{PGDs}$ model: early stopped at epoch 1105
    \item Avg+REx$_\textit{PGDs}$ model: REx penalty activated at epoch 551, early stopped at epoch 1105
    \item MSD model: early stopped at epoch 655
    \item MSD+REx model: REx penalty activated at epoch 101, early stopped at epoch 655
\end{itemize}

\textbf{CIFAR10}
\begin{itemize}
    \item $P_1$ model: early stopped at epoch 69
    \item $P_2$ model: early stopped at epoch 59
    \item $P_\infty$ model: early stopped at epoch 45
    \item Avg model: early stopped at epoch 50
    \item Avg+REx model: REx penalty activated at epoch 301, early stopped at epoch 330
    \item Avg$_\textit{PGDs}$ model: early stopped at epoch 95
    \item Avg+REx$_\textit{PGDs}$ model: REx penalty activated at epoch 301, early stopped at epoch 370
    \item MSD model: early stopped at epoch 40
    \item MSD+REx model: REx penalty activated at epoch 26, early stopped at epoch 70
\end{itemize}

\subsubsection{With hyperparameter optimisation} \label{apdx:subsec:with-hyperparameter-optimisation}
The experiments are run with a ResNet18 architecture on CIFAR10. We follow the results of \citet{rice2020overfitting} and \citet{pang2020bag}, using a piecewise learning rate schedule decay and a weight decay value of $5\cdot 10^{-4}$. In all cases we start with a learning rate of $0.1$, decayed to $0.01$ at the corresponding milestone. In preliminary tuning experiments, we observe the Avg$_\textit{PGDs}$ to perform very poorly relative to other baselines, and chose to drop that baseline.
\begin{itemize}
    \item $P_\infty$ model: milestone at epoch 100, early stopped at epoch 103
    \item Avg model: milestone at epoch 100, early stopped at epoch 140
    \item Avg+REx model: REx penalty activated at epoch 50 (before lr decay), milestone at epoch 100, early stopped at epoch 110
    \item MSD model: milestone at epoch 50, early stopped at epoch 51
    \item MSD+REx model: REx penalty activated at epoch 50 (before lr decay), milestone at epoch 97, early stopped at epoch 99
\end{itemize}
We attempt several learning rate schedule milestones for both MSD and MSD+REx, which has a higher impact than for other models. This process can be automated since the worst-case seen accuracy always peaks within a few epochs of a milestone.

\subsection{Other implementation details}
We use the implementation of \url{https://github.com/kuangliu/pytorch-cifar/blob/master/models/resnet.py} for ResNet18. For StAdv and RecolorAdv, we use the code of \citet{laidlaw2019functional}, available at \url{https://github.com/cassidylaidlaw/ReColorAdv}.

REx's $\beta$ parameter is generally set to $10$, except for MSD+REx on MNIST in subsection \ref{subsec:MNIST} where it is set to $4$. These numbers initially come from setting $\beta$ to a value of the same order of magnitude as $\frac{L_\textit{Avg}}{\Var}$'s value at the epoch REx is activated, in the early iterations of our experiments. This is done to encourage the optimisation dynamics to neglect neither term of the REx loss. We found that $\beta=10$ worked generally well, even in many settings where empirically $\frac{L_\textit{Avg}}{\Var} \simeq 30$. See Sec. \ref{apdx:sec:varying_beta} for a discussion of how the choice of $\beta$ affects performance.

\newpage
\section{Additional results}\label{apdx:sec:additional_results}
\subsection{More about the attack discriminator}\label{apdx:subsec:attack_discriminator}
\begin{wrapfigure}{r}{6cm}
\vspace{-0.4cm}
    \caption{Confusion matrix of ViT predicting whether perturbations stem from $P_\infty$ or $\textit{DF}_\infty^\bullet$ (normalised by row).}\label{fig:ViT_Pinf_DFinfmod}
    \includegraphics[width=5.5cm]{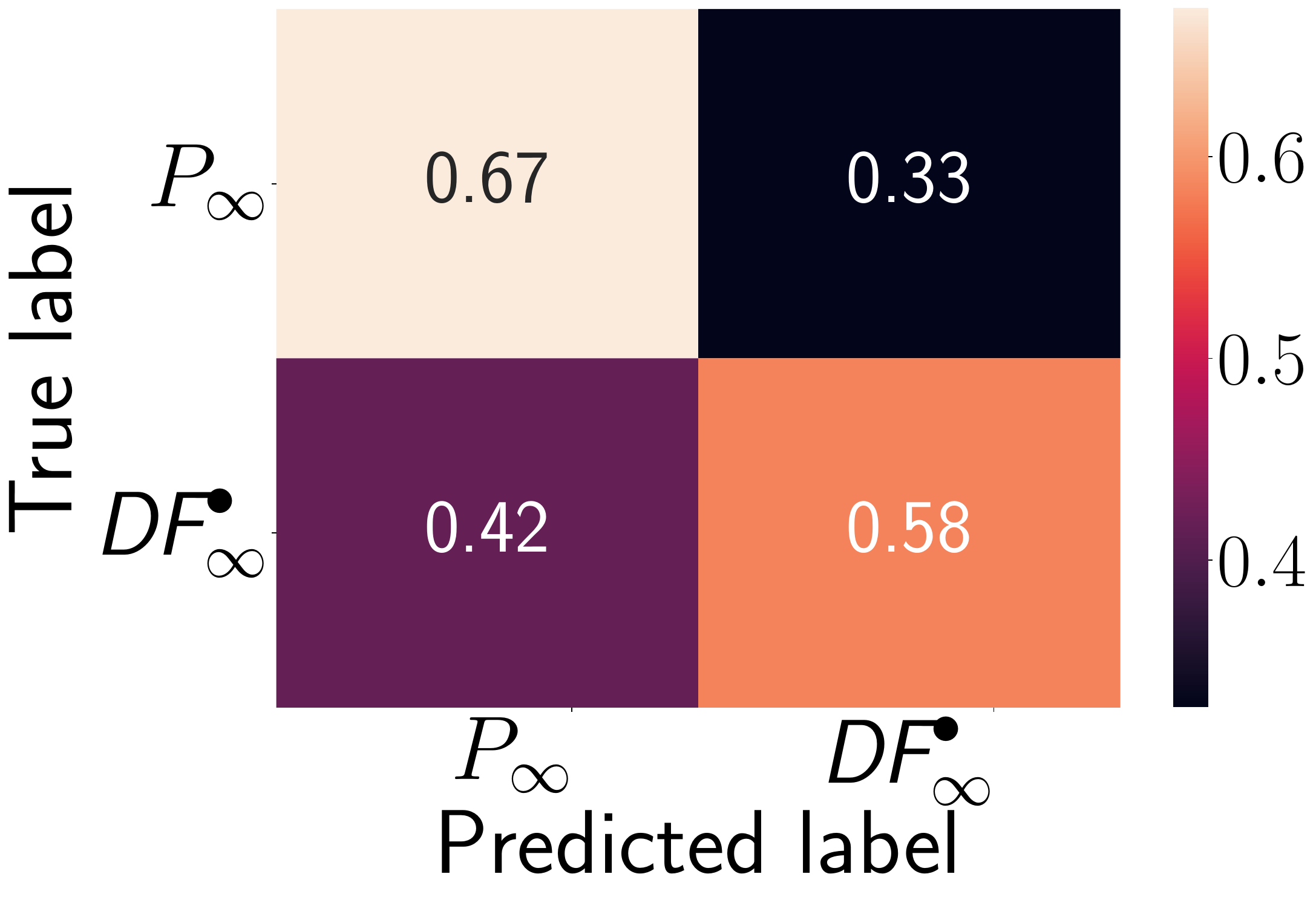}
\vspace{-0.3cm}
\end{wrapfigure}
We invite the reader to consult Apdx \ref{apdx:subsec:methodology_ViT_attack_discriminator} for details about the methodology of the results presented here and in subsection \ref{subsec:attack_discriminator}.

Here, we give more results about the attack discriminator. To claim unequivocally that all attacks considered are distributionally shifted with respect to one another, we require $p(A_\textit{true}) > p(A\neq A_\textit{true})$ where $A_\textit{true}$ is the true attack used to generate the sample and $p(A)$ is the probability predicted by the model that a perturbation corresponds to attack $A$. \footnote{While this is out of the scope of this work, it is possible to relate the divergence between the distributions to the performance of the discriminator, as done by \citet{bashivan2021adversarial} using \citet{kifer2004detecting}'s $\mathcal{H}\Delta \mathcal{H}$ divergence theory.} This is true for all attacks as seen in Fig. \ref{fig:ViT_discriminate_all_attacks}, except $\textit{DF}_\infty^\bullet$ being confused for $P_\infty$, and $\textit{CW}_2$ being confused for $\textit{CW}_2^\bullet$ or $\text{AutoAttack}_\infty$.

$\textit{DF}_\infty^\bullet$ and $P_\infty$ share the same $\epsilon=8/255$ and the model on which the attacks were generated (the $P_\infty$ adversarially trained model from Table \ref{table:tuned_cifar}) performs roughly equally on both (46.3\% accuracy on $\textit{DF}_\infty^\bullet$ vs 47.3\% on $P_\infty$). Therefore, it would be particularly important for our claim to be able to measure some difference between those distributions, even if there may be some overlap. Note that despite sharing a similar performance and $\epsilon$, $\text{AutoAttack}_\infty$ does not induce the same confusion.

In order to tackle this, we train a binary ViT classifier to tell apart $\textit{DF}_\infty^\bullet$ and $P_\infty$. We can see in Fig. \ref{fig:ViT_Pinf_DFinfmod} that the model is now able to properly predict each class with higher probability than the other, with average probability 63\%. The confusion (off-diagonal entries) are indicative that as we might suspect, there is \textit{some} overlap between the distributions of perturbations generated by the two attacks.

Concerning the $\textit{CW}_2$ confusion results, this is an artefact of the implementation of $\textit{CW}_2$ that we use. Indeed, the implementation of the attack from advertorch early stops by default if the iterations get stuck in a local minimum. This leads to some samples having near 0 perturbation and not being adversarial. As argued in the main text, this is no longer an issue as soon as $\textit{CW}_2$ is tuned so that this does not happen, e.g. with the $\textit{CW}_2^\bullet$ tuning.

\subsection{CIFAR10 - no hyperparameter optimisation}\label{apdx:subsec:CIFAR10}

\begin{table*}[h]
\centering
\caption{Accuracy on CIFAR10 for different domains. Ensembles omit $\textit{CW}_2^\bullet$ due to overtuning.}\label{table:cifar}
\resizebox{0.99\textwidth}{!} {
\begin{tabular}{@{}p{25mm}cccccccccc<{\kern-\tabcolsep}}
\hline
                                   & \multicolumn{10}{c}{Defenses}                                           \\ \hline
                                   & \text{None}                         & \multicolumn{3}{c}{Adversarial training}                                                   & \text{Avg}                          & \text{Avg+REx}                               & $\text{Avg}_{\textit{PGDs}}$                   & $\text{Avg+REx}_{\textit{PGDs}}$                        & \text{MSD}                          & \text{MSD+REx}                               \\ \hline
No attack                          & \cellcolor[HTML]{DDEBF7}87.6 & 92.1                         & 87.1                         & 77.4                         & \cellcolor[HTML]{DDEBF7}80.9 & \cellcolor[HTML]{DDEBF7}75.1          & \cellcolor[HTML]{DDEBF7}82.8 & \cellcolor[HTML]{DDEBF7}79.0          & 76.3 & \cellcolor[HTML]{DDEBF7}75.1          \\
$P_1$                              & 80.3                         & \cellcolor[HTML]{DDEBF7}87.6 & 85.0                         & 75.7                         & 78.7                         & 72.0                                  & \cellcolor[HTML]{DDEBF7}80.4 & \cellcolor[HTML]{DDEBF7}77.0          & \cellcolor[HTML]{DDEBF7}74.4 & \cellcolor[HTML]{DDEBF7}72.7          \\
$P_2$                              & 19.9                         & 47.8                         & \cellcolor[HTML]{DDEBF7}70.9 & 66.6                         & 69.8                         & 65.3                                  & \cellcolor[HTML]{DDEBF7}71.1 & \cellcolor[HTML]{DDEBF7}68.0          & \cellcolor[HTML]{DDEBF7}65.7 & \cellcolor[HTML]{DDEBF7}65.9          \\
$P_\infty$                         & 0.0                          & 0.0                          & 9.7                          & \cellcolor[HTML]{DDEBF7}39.5 & \cellcolor[HTML]{DDEBF7}34.4 & \cellcolor[HTML]{DDEBF7}\textbf{44.2} & \cellcolor[HTML]{DDEBF7}32.3 & \cellcolor[HTML]{DDEBF7}\textbf{41.2} & \cellcolor[HTML]{DDEBF7}37.9 & \cellcolor[HTML]{DDEBF7}\textbf{41.9} \\
$DF_\infty$                        & 4.1                          & 19.2                         & 60.2                         & 64.6                         & \cellcolor[HTML]{DDEBF7}64.9 & \cellcolor[HTML]{DDEBF7}62.2          & 66.8                         & 64.5                                  & 62.7                         & 63.6                                  \\
$CW_2$                             & 0.0                          & 0.0                          & 1.3                          & 11.2                         & \cellcolor[HTML]{DDEBF7}9.8  & \cellcolor[HTML]{DDEBF7}\textbf{21.6} & 8.7                          & \textbf{16.5}                         & 10.7                         & \textbf{17.5}                         \\
$P_\infty^\bullet$                         & 0.0                          & 0.0                          & 1.0                          & 20.1                         & 16.3                         & \textbf{24.1}                         & 13.2                         & \textbf{22.1}                         & 19.3                         & \textbf{23.8}                         \\
$DF_\infty^\bullet$                        & 0.0                          & 0.0                          & 9.5                          & 38.5                         & 35.6                         & \textbf{40.5}                         & 33.0                         & \textbf{39.1}                         & 36.6                         & \textbf{40.4}                         \\
AutoAttack$_\infty$                     & 0.0                          & 0.0                          & 8.1                          & 37.2                         & 33.7                         & \textbf{38.8}                         & 31.2                         & \textbf{37.6}                         & 36.0                         & \textbf{39.0}                         \\ \hline
Ensemble (seen)                    & -                            & -                            & -                            & -                            & 9.8                          & \textbf{21.2}                         & 32.3                         & \textbf{41.2}                         & 37.9                         & \textbf{41.8}                         \\
Ensemble (unseen by all models) & 0.0                          & 0.0                          & 1.0                          & 20.1                         & 16.3                         & \textbf{24.0}                         & 13.2                         & \textbf{22.0}                         & 19.3                         & \textbf{23.6}                         \\
Ensemble (unseen by this model)    & 0.0                          & 0.0                          & 0.9                          & 10.7                         & 16.3                         & \textbf{24.0}                         & 7.7                          & \textbf{14.2}                         & 10.3                         & \textbf{16.1}                         \\
Ensemble (all)           & 0.0                          & 0.0                          & 0.9                          & 10.7                         & 9.4                          & \textbf{17.9}                         & 7.7                          & \textbf{14.2}                         & 10.3                         & \textbf{16.1}                         \\ \hline
$CW_2^\bullet$                             & 0.0                          & 0.0                          & 0.1                          & 1.1                          & 1.6                          & \textbf{2.6}                          & 1.1                          & \textbf{2.7}                          & 1.0                          & \textbf{2.7}                          \\ \hline
\end{tabular}
}
\end{table*}
The results on CIFAR10 without performing hyperparameter optimisation on each model are summarised in Table \ref{table:cifar}. We observe again that REx is an improvement over the ensemble of seen attacks compared to the baselines it was used on. As on MNIST, this happens by improving the performance on the strongest of the seen attacks and sacrificing a little performance on the top performing attack(s). Moreover, REx consistently yields a significant improvement in robustness when evaluated against the ensemble of unseen attacks, too. The only individual domains where REx never yields significant improvements are the clean (unperturbed) data, $P_1$ and $P_2$, whether they were seen during training or not. Given the relatively good performance of the baselines and REx on those domains, this is in line with REx's tendency to sacrifice a little accuracy on the best performing domains to improve significantly the performance on the worst performing domains. 

While adversarial training on either $P_1$ or $P_2$ fails to yield robustness to unseen attacks, we observe that these two defenses are the only ones for which the clean accuracy does not decrease significantly. We note that unlike on MNIST, MSD is significantly more competitive with the other baselines, and its performance is relatively similar to the one reported by \citet{maini2020adversarial} (likely due to using the same architecture on CIFAR10).

As with tuned models (subsec. \ref{subsec:tuned_CIFAR10}), the model trained on $P_\infty$ performs better than the Avg, Avg$_\textit{PGDs}$ and MSD models on the set of attacks unseen by all models. 
Conversely, training on ensembles of attacks also hurts performance on $P_\infty$, unless we apply REx.
In other words, only REx is able to improve both $P_\infty$ \textit{and} worst-case performance over an ensemble of attacks.
Moreover, without hyperparameter tuning, REx appears to lose more performance on the best performing domains. This is particularly notable in the case of $P_2$ attacks, where for example REx improves the $P_2$ accuracy by $+0.8$\% with hyperparameter tuning, vs $-4.5$\% without with the Avg model training on $\{P_\infty, \textit{DF}_\infty, \textit{CW}_2\}$. In contrast, the gap in performance in performance between MSD and MSD+REx is even larger than without individual hyperparameter optimisation, showing that REx helps bridge gaps in performance due to suboptimal tuning.

\begin{figure*}[h]

  \begin{subfigure}[t]{.49\textwidth}
    \centering
    \includegraphics[width=\linewidth]{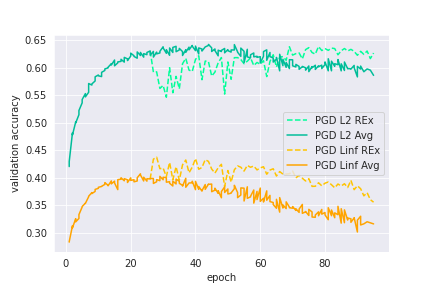}
    \caption{CIFAR10 seen attacks}
  \end{subfigure}
  \hfill
  \begin{subfigure}[t]{.49\textwidth}
    \centering
    \includegraphics[width=\linewidth]{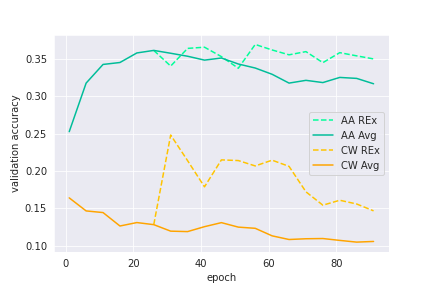}
    \caption{CIFAR10 unseen attacks}
  \end{subfigure}
  \caption{Validation accuracy of MSD on CIFAR10 (bottom)
  with and without REx (dashed line), against seen attacks (left) and unseen attacks (right) (AA=AutoAttack).}\label{fig:validation_with_REx_CIFAR}
\end{figure*}
As with MNIST and the CIFAR10 hyperparameter-optimised models, early stopping is important, and moreso than with MNIST, the performance of REx performance quickly peaks, as seen in Fig. \ref{fig:validation_with_REx_CIFAR}.

\begin{tcolorbox}[leftrule=1.5mm,top=0.8mm,bottom=0.5mm]
\textbf{Key observations 4 (no hyperparameter tuning):}
\begin{itemize}
    \item REx improves the performance of all baselines on CIFAR10 with a ResNet18, from 10.7\% with the best baseline to 17.9\% accuracy against an ensemble of attacks, by sacrificing a little robustness against the weakest individual attacks.
    \item Multi-perturbation defenses only achieve higher $P_\infty$ and worst-case performance than $P_\infty$ adversarial training when using REx.
    \item REx helps bridge gaps in performance due to suboptimal hyperparameter tunings.
\end{itemize}
\end{tcolorbox}

\subsection{Perceptual Adversarial Training and additional experiments}
\textbf{Perceptual Adversarial Training}: due to the similarity of our motivation with that of Perceptual Adversarial Training (PAT) \citep{laidlaw2020perceptual} (i.e. be robust against imperceptible attacks), we evaluate their model on our benchmark. We simply run the code made publicly available by the authors in order to train with PAT a model, and load it in our evaluation pipeline. With a ResNet18 on CIFAR10, we find that they achieve considerably worse robustness, with Ensemble (all) worst-case robustness of $3.8\%$. The model is noticeably weak against $L_\infty$ attacks, achieving $6.9\%$ accuracy on AutoAttack and $10.0\%$ on $P_\infty$. While this may be surprising, this can be explained by noting that \citet{laidlaw2020perceptual} use a ResNet50 on CIFAR10, while we used their code with the "--arch resnet18" argument to train a model immediately comparable to the main results of our work.

\textbf{Robustness against different tunings of AutoAttack}: we consider an additional tuning of AutoAttack $L_\infty$ on CIFAR10 with the hyperparameter-optimised ResNet18, this time with $\epsilon=12/255$. We find that the MSD model achieves $25.0\%$ accuracy while MSD+REx achieves $26.8\%$. The Avg model from Table \ref{table:tuned_cifar} achieves $15.6\%$ accuracy while its REx counterpart reaches $24.2\%$ in 30 less epochs of training (see Sec. \ref{apdx:subsec:with-hyperparameter-optimisation} for number of epochs for each model). The $P_\infty$ model yields $25.9\%$, which is higher than the other baselines, but still worse than MSD+REx. This further supports the claim that REx improves robustness against unseen attacks and tunings of attacks.

\subsection{Additional results on CIFAR10-C}
\begin{table}[h]
\centering
\caption[width=0.5\textwidth]{Accuracies of tuned CIFAR10 models on CIFAR10-C corruptions.}\label{table:cifar-c}
\begin{tabular}{@{}p{25mm}cccccc<{\kern-\tabcolsep}}
\hline
                                   & \multicolumn{6}{c}{Defenses}                                           \\ \hline
                   & None & $P_\infty$ & Avg & Avg+REx & MSD & MSD+REx \\
                   \hline
brightness         & 31.7                     & 61.5                     & 29.1                    & 54.5                    & 51.7                   & 60.2                   \\
contrast           & 35.4                     & 43.7                     & 33.4                    & 40.6                    & 39.8                   & 42.7                   \\
defocus blur      & 22.9                     & 52.6                     & 25.8                    & 49.1                    & 44.9                   & 52.8                   \\
elastic transform & 21.6                     & 50.1                     & 24.5                    & 46.4                    & 42.1                   & 50.3                   \\
fog                & 32.5                     & 45.8                     & 32.7                    & 44.2                    & 41.7                   & 45.8                   \\
frost              & 28.4                     & 63.1                     & 33.6                    & 56.8                    & 54.5                   & 61.0                   \\
gaussian blur     & 22.2                     & 51.4                     & 25.7                    & 48.1                    & 44.5                   & 52.0                   \\
gaussian noise    & 13.4                     & 53.3                     & 25.1                    & 48.7                    & 35.4                   & 50.9                   \\
glass blur        & 18.0                     & 49.7                     & 24.2                    & 46.1                    & 39.8                   & 48.3                   \\
impulse noise     & 14.7                     & 47.7                     & 24.8                    & 45.6                    & 33.0                   & 46.0                   \\
jpeg compression  & 18.6                     & 54.7                     & 25.4                    & 50.5                    & 44.8                   & 53.7                   \\
motion blur       & 23.2                     & 49.6                     & 24.9                    & 46.3                    & 42.9                   & 50.4                   \\
pixelate           & 20.9                     & 54.8                     & 25.3                    & 50.2                    & 44.6                   & 53.2                   \\
saturate           & 20.8                     & 49.1                     & 20.8                    & 43.6                    & 37.6                   & 47.6                   \\
shot noise        & 13.6                     & 52.3                     & 24.8                    & 48.4                    & 34.5                   & 50.3                   \\
snow               & 22.3                     & 57.9                     & 27.6                    & 52.4                    & 47.9                   & 55.8                   \\
spatter            & 18.4                     & 52.6                     & 24.6                    & 48.4                    & 42.4                   & 50.5                   \\
speckle noise     & 13.2                     & 50.9                     & 24.5                    & 47.5                    & 32.3                   & 49.0                   \\
zoom blur         & 22.3                     & 52.3                     & 26.1                    & 48.1                    & 45.2                   & 52.7                   \\ \hline
average            & 21.8                     & 52.3                     & 26.5                    & 48.2                    & 42.1                   & 51.2        \\ \hline          
\end{tabular}
\end{table}
In order to test the investigate the robustness of the different tuned baselines against non-adversarial perturbations, we evaluate them on the corruptions of CIFAR10-C \citep{hendrycks2019robustness}. Table \ref{table:cifar-c} shows how $P_\infty$ is the overall best performing model. The Avg and MSD baselines perform surprisingly poorly on CIFAR10-C, and REx leads to very sigificant improvements on both baselines that bring the performance closer to that of the model trained on $P_\infty$. In particular, MSD+REx comes very close (within $2\%$ accuracy) to the $P_\infty$ baseline's performance, only outperforming $P_\infty$ marginally on defocus blurs, elastic transforms, gaussian blurs, motion blurs, and zoom blurs, and having equal performance on fog. This is in contrast with the intuition that because REx had lower (in-distribution) clean accuracy than most other baselines on CIFAR10, this would also be true for out-of-distribution non-adversarial data.

\subsection{Transferability of REx models}
Several works have highlighted how adversarial robustness can be of interest for transfer learning \citep{salman2020adversarially, utreraadversarially}. We freeze the parameters of the hyperparameter-optimised CIFAR10 models, only resetting the last linear layer (and adapting it to the appropriate number of output classes) and allowing it to train on the CIFAR100 \citep{krizhevsky2009learning} and SVHN \citep{netzer2011reading} training datasets, then evaluating them on their test sets. We train for 30 epochs with a learning rate of 0.1, decaying by factors 0.1 at epochs 10 and 20. We repeat this 3 times per model, using a different seed ("0", "1" and "2") to report average accuracies on the test set and standard deviations. 

\textbf{Disclaimer:} we did not attempt any particular optimisation of these results with state of the art techniques, and merely provide these results for completeness. We also highlight that work on transfer learning with robust models often focuses on larger models (e.g. ResNet50 for \citet{utreraadversarially}), which can significantly affect the results. Finally, we realise that transferring from CIFAR10 to CIFAR100 (instead of the other way around) may be overly ambitious, with works such as \citep{utreraadversarially} focusing instead on CIFAR100 to CIFAR10.

\begin{table}[h]
\centering
\caption{Accuracies of tuned CIFAR10 models finetuned on CIFAR100, averaged over 3 finetunings with different seeds per defense.}\label{table:cifar_finetuned_to_CIFAR100}
\begin{tabular}{@{}p{25mm}cccccc<{\kern-\tabcolsep}}
\hline
                                   & \multicolumn{6}{c}{Defenses}                                           \\ \hline
                   & None & $P_\infty$ & Avg & Avg+REx & MSD & MSD+REx \\
                   \hline
Accuracy & 28.9$\pm$0.1                 & 29.3$\pm$0.1                 & 25.9$\pm$0.2                & 27.7$\pm$0.2                & 28.4$\pm$0.1               & 27.7$\pm$0.1           \\ \hline
\end{tabular}
\end{table}
On CIFAR100, we can see on Table \ref{table:cifar_finetuned_to_CIFAR100} that the $P_\infty$ model transfers best, marginally better than the non-robust model. All other baseline perform worse than the non-robust model, and REx only appear to improve the Avg baseline, while decreasing by $0.7\%$ the accuracy when used on the MSD baseline before the finetuning. The performance of the non-robust model is perhaps not surprising, given the similarity between the CIFAR10 and CIFAR100 classes, which may explain the relative performances of the baselines being correlated with those observed on CIFAR10 on unperturbed data.

\begin{table}[h]
\centering
\caption{Accuracies of tuned CIFAR10 models finetuned on SVHN, averaged over 3 finetunings with different seeds per defense.}\label{table:cifar_finetuned_to_SVHN}
\begin{tabular}{@{}p{25mm}cccccc<{\kern-\tabcolsep}}
\hline
                                   & \multicolumn{6}{c}{Defenses}                                           \\ \hline
                   & None & $P_\infty$ & Avg & Avg+REx & MSD & MSD+REx \\
                   \hline
Accuracy & 42.4$\pm$0.2                 & 49.2$\pm$0.3                 & 44.8$\pm$0.1                & 47.1$\pm$0.1                & 49.1$\pm$0.2               & 48.0$\pm$0.1              \\ \hline
\end{tabular}
\end{table}
In order to investigate this hypothesis, we also attempt a similar finetuning on the SVHN dataset. The (cropped) SVHN dataset consists in pictures of house numbers, where the task is to read the digit at the center of the image. Therefore, this task corresponds to a shift substantially different than CIFAR100. As suspected, we observe in Table \ref{table:cifar_finetuned_to_SVHN} than all of the robust models now have significantly better performance than the non-robust model. The $P_\infty$ and MSD models both transfer the best, with the MSD+REx model losing again roughly $1\%$ of performance compared to MSD, but still ahead of the non-robust model or the Avg and Avg+REx models.

\newpage
\section{When to early stop, and how learning rate milestones affect performance}\label{apdx:sec:when_to_early_stop}
This section shows results about learning rate schedule milestones and early stopping for the CIFAR10 models with hyperparameter optimisation.
\subsection{MSD model}
\vspace{-0.2\baselineskip}
\begin{figure*}[h]
  \begin{subfigure}[t]{.45\textwidth}
    \centering
    \includegraphics[width=\linewidth]{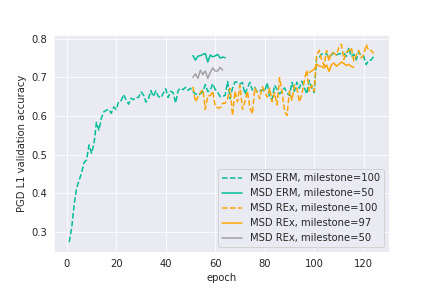}
    \caption{$P_1$ accuracy}
  \end{subfigure}
  \hfill
  \begin{subfigure}[t]{.45\textwidth}
    \centering
    \includegraphics[width=\linewidth]{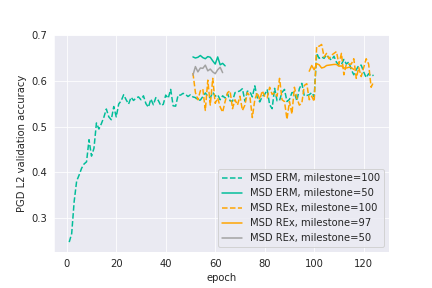}
    \caption{$P_2$ accuracy}
  \end{subfigure}

\vspace{-0.2\baselineskip}
  \begin{subfigure}[t]{.45\textwidth}
    \centering
    \includegraphics[width=\linewidth]{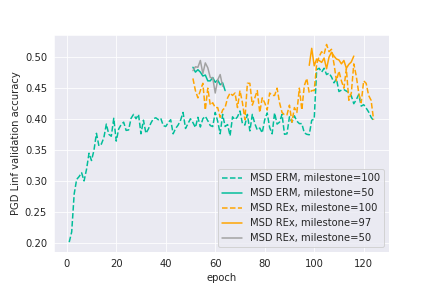}
    \caption{$P_\infty$ accuracy}
  \end{subfigure}
  \hfill
  \begin{subfigure}[t]{.45\textwidth}
    \centering
    \includegraphics[width=\linewidth]{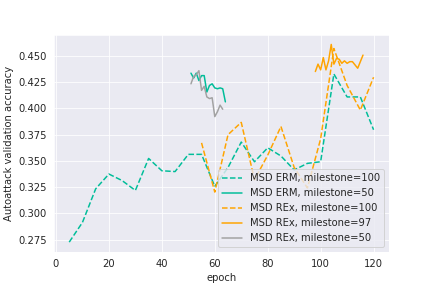}
    \caption{AutoAttack$_\infty$ accuracy}
  \end{subfigure}

\vspace{-0.2\baselineskip}
  \begin{subfigure}[t]{.45\textwidth}
    \centering
    \includegraphics[width=\linewidth]{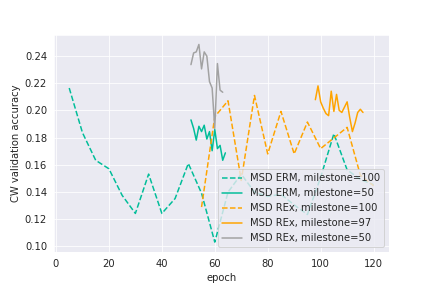}
    \caption{CW$_2$ accuracy}
  \end{subfigure}
  \hfill
  \begin{subfigure}[t]{.45\textwidth}
    \centering
    \includegraphics[width=\linewidth]{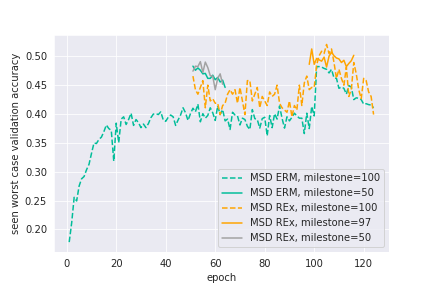}
    \caption{Ensemble (seen) accuracy}
  \end{subfigure}

\vspace{-0.2\baselineskip}
  \begin{subfigure}[t]{.45\textwidth}
    \centering
    \includegraphics[width=\linewidth]{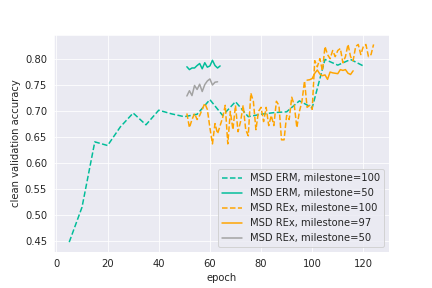}
    \caption{No attack (clean) accuracy}
  \end{subfigure}
  \hfill
  \begin{subfigure}[t]{.45\textwidth}
    \centering
    \includegraphics[width=\linewidth]{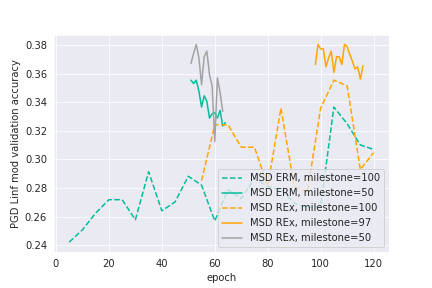}
    \caption{$P_\infty^\bullet$ accuracy}
  \end{subfigure}
  \caption{Validation accuracy of MSD and MSD+REx on CIFAR10 on various attacks with different milestones for the learning rate decay. Early stopping is performed for each model at the peak of the ensemble (seen) accuracy. The MSD model with a milestone at epoch 50 and the MSD+REx model with a milestone at epoch 97 are the final models retained in subsection \ref{subsec:tuned_CIFAR10}.}\label{fig:when_to_early_stop_MSD}
\end{figure*}
\vspace{-0.2\baselineskip}
In this subsection, we illustrate two points using Fig. \ref{fig:when_to_early_stop_MSD}: how early stopping is performed on the MSD and MSD+REx models, and how the choice of learning rate decay milestone affects performance. Regarding the former, we early stop based on the peak of the validation Ensemble (seen) accuracy. We motivated that worst-case performance is a more appropriate notion of robustness, and unseen attacks should not be used to make model-selection decisions as they are used to simulate ``future'', novel attacks that were not known when designing the defenses.
Concerning early stopping, we note that the peak performance on Ensemble (seen) accuracy is reached:
\begin{itemize}
    \item At epoch 101 for the MSD model with milestone at epoch 100
    \item At epoch 51 for the MSD model with milestone at epoch 50
    \item At epoch 105 for the MSD+REx model with milestone at epoch 100
    \item At epoch 99 for the MSD+REx model with milestone at epoch 97
    \item At epoch 54 or 56 for the MSD+REx model with milestone at epoch 50.
\end{itemize}
Regarding how milestone choice affects performance, this illustrates how we searched for hyperparameters. We attempt learning rate decays milestones at various epochs, which we then evaluate only for a few epochs, as the performance decays fast anyway as predicted by \citet{rice2020overfitting}. The curves in Fig. \ref{fig:when_to_early_stop_MSD} represent the best learning rate scheduler milestones found for MSD and MSD+REx. We retain the model with best validation Ensemble (seen) accuracy for our final results presented in Sec. \ref{sec:results}. As performance of the best checkpoints of MSD with milestones 50 and 100, and respectively MSD+REx with milestones 97 and 100, are very close, in both cases we evaluated the final models on the test set and kept the best in each case (MSD with milestone 50 and MSD+REx with milestone 97), observing only minor differences between each defense's pair of choices.

\subsection{Avg model}
\begin{figure*}[h]
  \begin{subfigure}[t]{.45\textwidth}
    \centering
    \includegraphics[width=\linewidth]{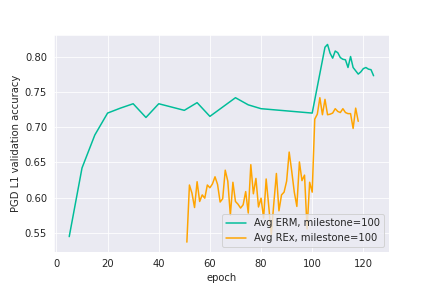}
    \caption{$P_1$ accuracy}
  \end{subfigure}
  \hfill
  \begin{subfigure}[t]{.45\textwidth}
    \centering
    \includegraphics[width=\linewidth]{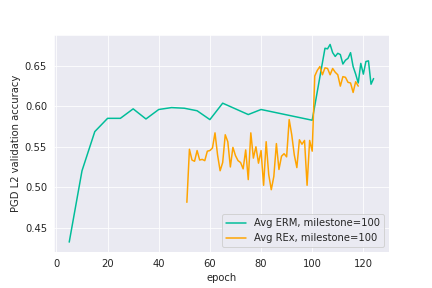}
    \caption{$P_2$ accuracy}
  \end{subfigure}

\vspace{-0.2\baselineskip}
  \begin{subfigure}[t]{.45\textwidth}
    \centering
    \includegraphics[width=\linewidth]{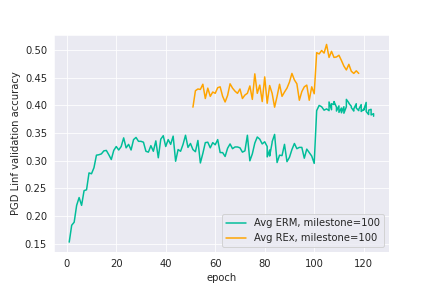}
    \caption{$P_\infty$ accuracy}
  \end{subfigure}
  \hfill
  \begin{subfigure}[t]{.45\textwidth}
    \centering
    \includegraphics[width=\linewidth]{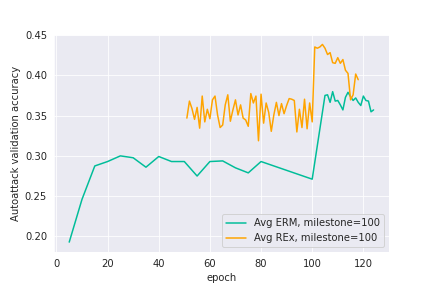}
    \caption{AutoAttack$_\infty$ accuracy}
  \end{subfigure}

\vspace{-0.2\baselineskip}
  \begin{subfigure}[t]{.45\textwidth}
    \centering
    \includegraphics[width=\linewidth]{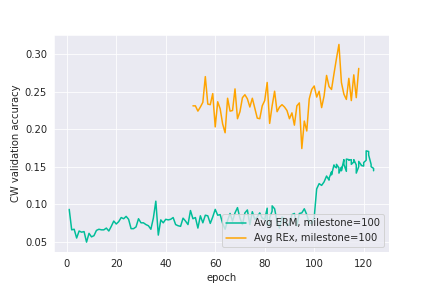}
    \caption{CW$_2$ accuracy}
  \end{subfigure}
  \hfill
  \begin{subfigure}[t]{.45\textwidth}
    \centering
    \includegraphics[width=\linewidth]{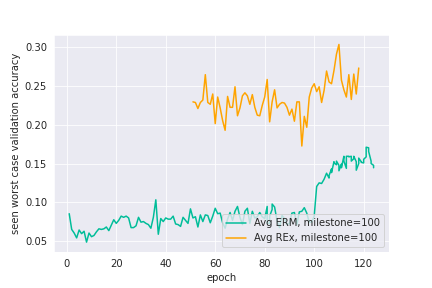}
    \caption{Ensemble (seen) accuracy}
  \end{subfigure}

\vspace{-0.2\baselineskip}
  \begin{subfigure}[t]{.45\textwidth}
    \centering
    \includegraphics[width=\linewidth]{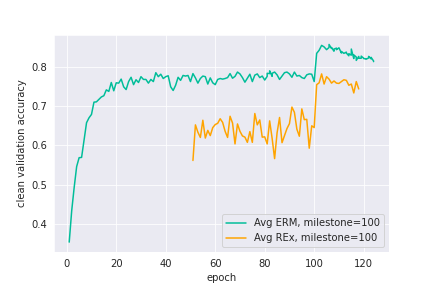}
    \caption{No attack (clean) accuracy}
  \end{subfigure}
  \hfill
  \begin{subfigure}[t]{.45\textwidth}
    \centering
    \includegraphics[width=\linewidth]{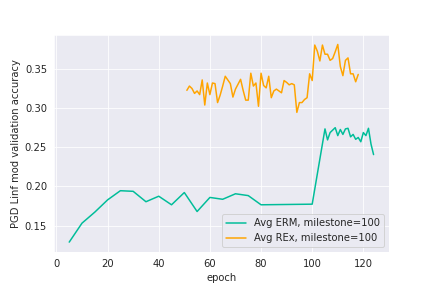}
    \caption{$P_\infty^\bullet$ accuracy}
  \end{subfigure}
  \caption{Validation accuracy of Avg and Avg+REx on CIFAR10 on various attacks with learning rate decay milestone at epoch 100. This illustrates how much easier it is to get improvements with REx on baselines based on ERM over multiple domains.}\label{fig:when_to_early_stop_Avg}
\end{figure*}

As argued in our introduction, it is somewhat surprising that REx successfully improved MSD due to MSD being a single-domain baseline. REx was originally designed to be used with baselines where multiple domains appear in the loss, and in particular ERM over multiple domains. Fig. \ref{fig:when_to_early_stop_Avg} illustrates how REx clearly benefits the Avg baseline more, with very little tuning effort required to achieve results above all baselines as reported in subsection \ref{subsec:tuned_CIFAR10}.

\vspace{-0.2\baselineskip}
\vspace{-0.2\baselineskip}

\newpage
\section{Effect of varying REx's $\beta$ coefficient}\label{apdx:sec:varying_beta}
\vspace{-0.2\baselineskip}
\begin{figure*}[h]
  \begin{subfigure}[t]{.48\textwidth}
    \centering
    \includegraphics[width=0.95\linewidth]{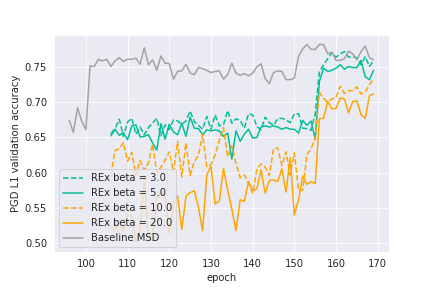}
    \caption{$P_1$ accuracy}
  \end{subfigure}
  \hfill
  \begin{subfigure}[t]{.48\textwidth}
    \centering
    \includegraphics[width=0.95\linewidth]{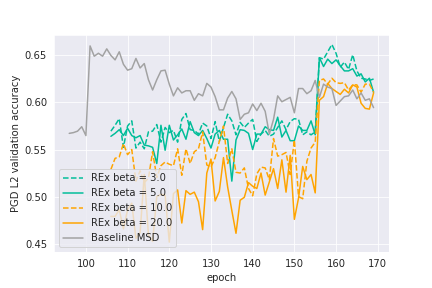}
    \caption{$P_2$ accuracy}
  \end{subfigure}

\vspace{-0.2\baselineskip}
  \begin{subfigure}[t]{.48\textwidth}
    \centering
    \includegraphics[width=0.95\linewidth]{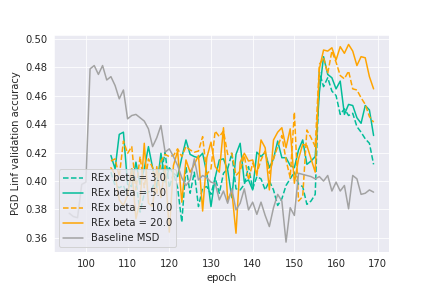}
    \caption{$P_\infty$ accuracy}
  \end{subfigure}
  \hfill
  \begin{subfigure}[t]{.48\textwidth}
    \centering
    \includegraphics[width=0.95\linewidth]{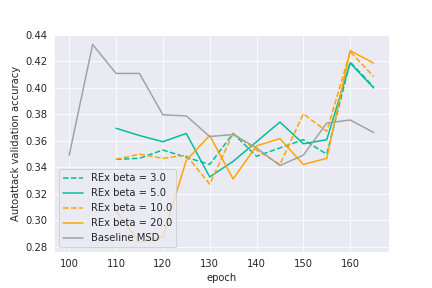}
    \caption{AutoAttack$_\infty$ accuracy}
  \end{subfigure}

\vspace{-0.2\baselineskip}
  \begin{subfigure}[t]{.48\textwidth}
    \centering
    \includegraphics[width=0.95\linewidth]{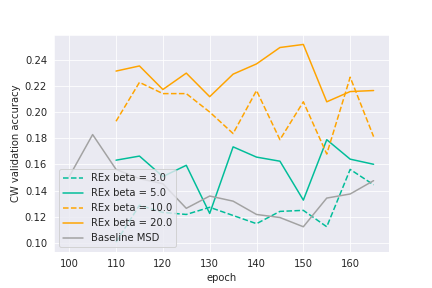}
    \caption{CW$_2$ accuracy}
  \end{subfigure}
  \hfill
  \begin{subfigure}[t]{.48\textwidth}
    \centering
    \includegraphics[width=0.95\linewidth]{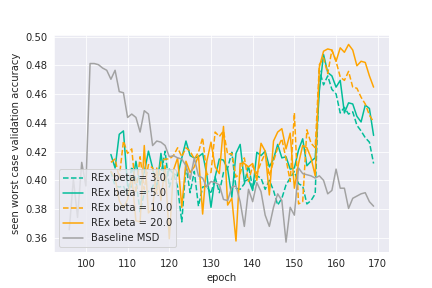}
    \caption{Ensemble (seen) accuracy}
  \end{subfigure}

\vspace{-0.2\baselineskip}
  \begin{subfigure}[t]{.48\textwidth}
    \centering
    \includegraphics[width=0.95\linewidth]{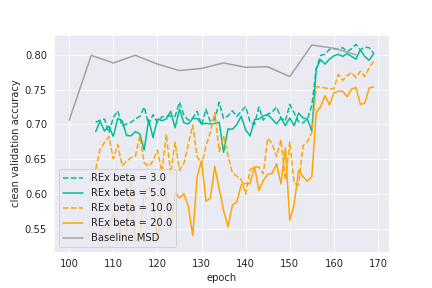}
    \caption{No attack (clean) accuracy}
  \end{subfigure}
  \hfill
  \begin{subfigure}[t]{.48\textwidth}
    \centering
    \includegraphics[width=0.95\linewidth]{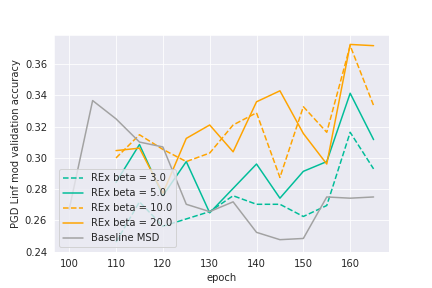}
    \caption{$P_\infty^\bullet$ accuracy}
  \end{subfigure}
  \caption{Validation accuracy of MSD and MSD+REx on CIFAR10 on various attacks with different values of $\beta$. Note that MSD has a learning rate decay at epoch 100, and MSD+REx at epoch 155.}\label{fig:varying_betas}
\end{figure*}
\vspace{-0.2\baselineskip}
In Fig. \ref{fig:varying_betas} we investigate the effect of varying $\beta$ in REx+MSD. Note that the hyperparameters of both MSD and MSD+REx are suboptimally tuned in this figure, which only aims to illustrate the effect of varying $\beta$. To generate those figures, both baselines use weight decay, MSD's learning rate milestone is set at epoch 100. MSD+REx loads an MSD checkpoint at epoch 105 with a learning rate set to 0.1 which is decayed at epoch 155.

We observe that while there is some robustness to the choice of $\beta$ for some domains, the difference is especially large on CW$_2$ and $P_\infty^\bullet$, where a larger value benefits the model. $\beta$ also has a fairly large impact on the clean, $P_1$ and $P_2$ accuracies. This is explained by the fact that low values of $\beta$ imply that the variance term will have lower impact and the model will value high performing seen domains (clean, $P_1$, $P_2$) more when updating weights than if larger values of $\beta$ were used. In contrast, large values of $\beta$ emphasise the variance regularisation which benefits accuracy against stronger attacks more.

\newpage
\section{Miscellaneous results}
\begin{figure*}[h]
  \begin{subfigure}[t]{\textwidth}
    \centering
    \includegraphics[width=0.85\linewidth]{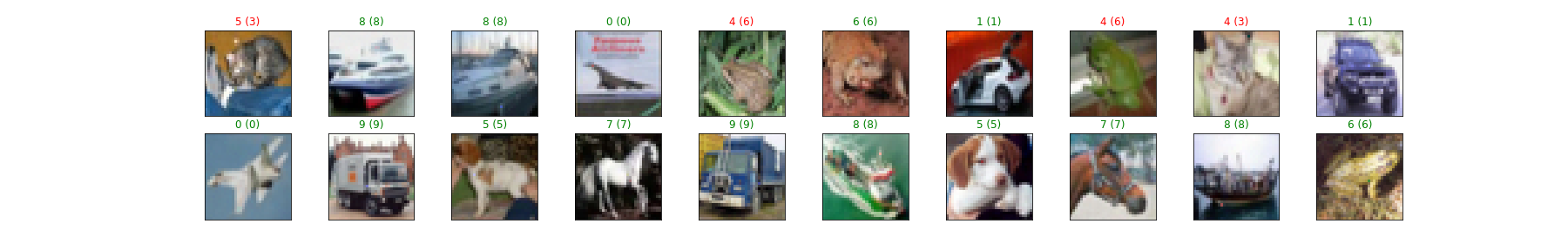}
    \caption{$P_1$ adversarial examples}
  \end{subfigure}
  \hfill
  \begin{subfigure}[t]{\textwidth}
    \centering
    \includegraphics[width=0.85\linewidth]{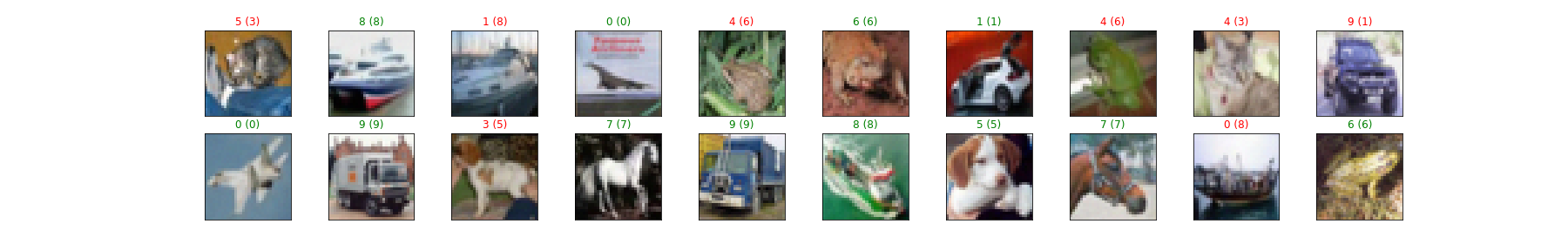}
    \caption{$P_2$ adversarial examples}
  \end{subfigure}

\vspace{-0.2\baselineskip}
  \begin{subfigure}[t]{\textwidth}
    \centering
    \includegraphics[width=0.85\linewidth]{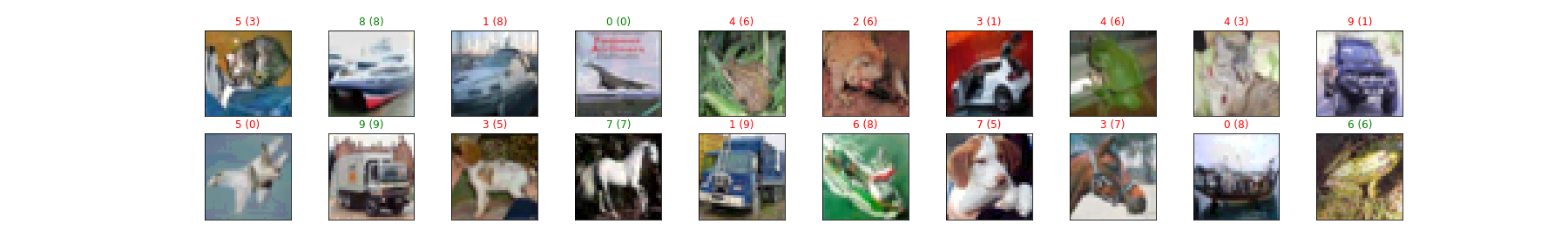}
    \caption{$P_\infty$ adversarial examples}
  \end{subfigure}
  \hfill
  \begin{subfigure}[t]{\textwidth}
    \centering
    \includegraphics[width=0.85\linewidth]{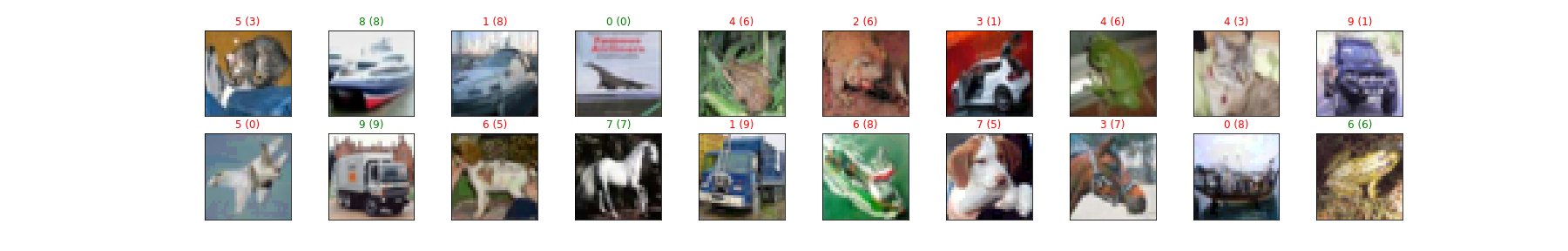}
    \caption{AutoAttack$_\infty$ adversarial examples}
  \end{subfigure}

\vspace{-0.2\baselineskip}
  \begin{subfigure}[t]{\textwidth}
    \centering
    \includegraphics[width=0.85\linewidth]{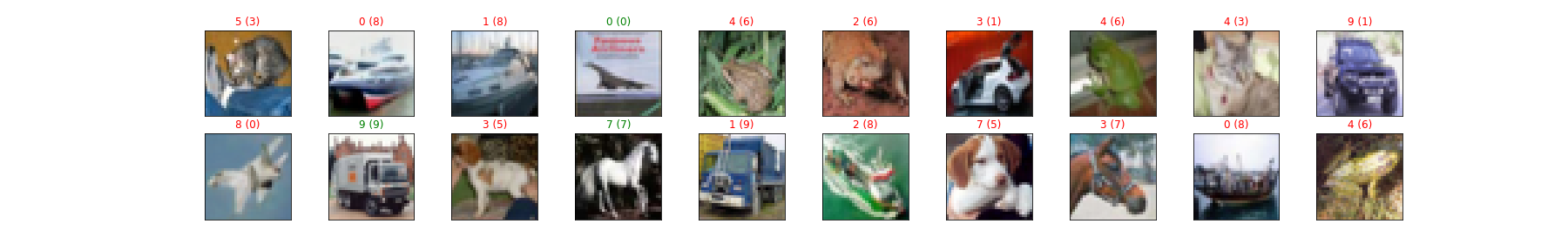}
    \caption{CW$_2$ adversarial examples}
  \end{subfigure}
  \hfill
  \begin{subfigure}[t]{\textwidth}
    \centering
    \includegraphics[width=0.85\linewidth]{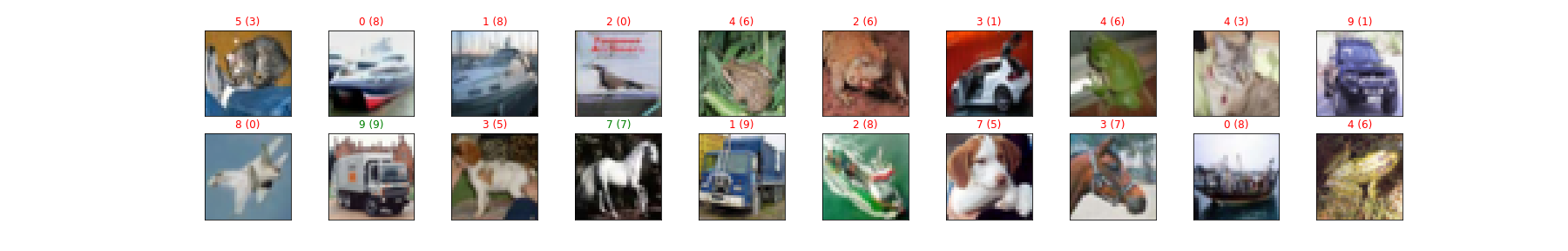}
    \caption{CW$_2^\bullet$ examples}
  \end{subfigure}

\vspace{-0.2\baselineskip}
  \begin{subfigure}[t]{\textwidth}
    \centering
    \includegraphics[width=0.85\linewidth]{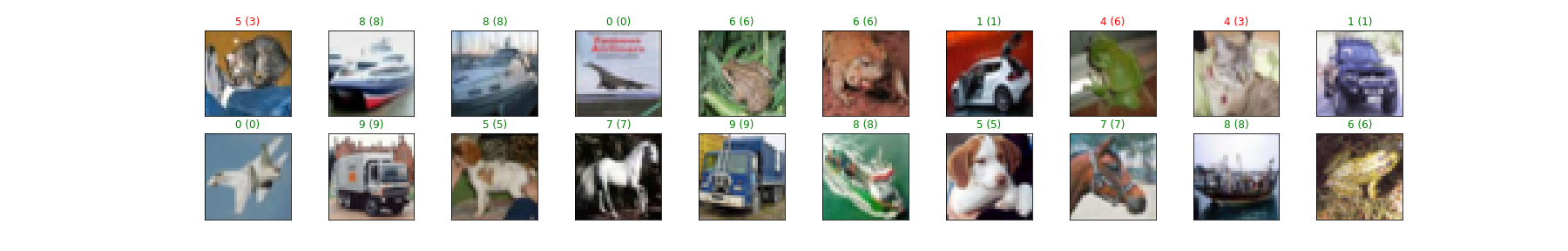}
    \caption{No attack (clean) adversarial examples}
  \end{subfigure}
  \hfill
  \begin{subfigure}[t]{\textwidth}
    \centering
    \includegraphics[width=0.85\linewidth]{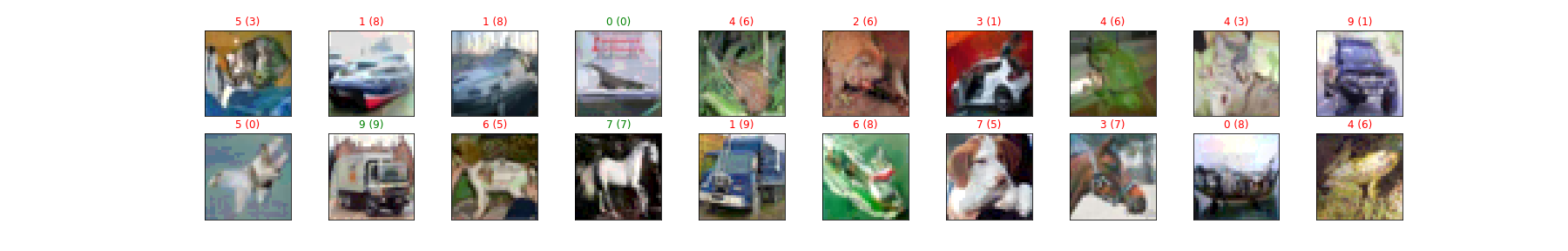}
    \caption{$P_\infty^\bullet$ adversarial examples}
  \end{subfigure}
  \caption{Adversarial examples generated from the hyperparameter-optimised Avg model, for each attack. The predicted class and in parentheses, the true class, are displayed above each image.}\label{fig:adv_examples_Avg}
\end{figure*}

\begin{figure*}[h]
  \begin{subfigure}[t]{\textwidth}
    \centering
    \includegraphics[width=0.85\linewidth]{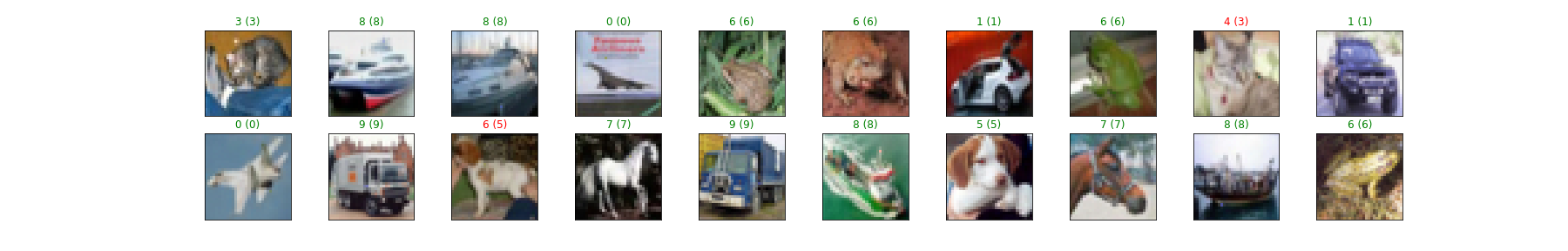}
    \caption{$P_1$ adversarial examples}
  \end{subfigure}
  \hfill
  \begin{subfigure}[t]{\textwidth}
    \centering
    \includegraphics[width=0.85\linewidth]{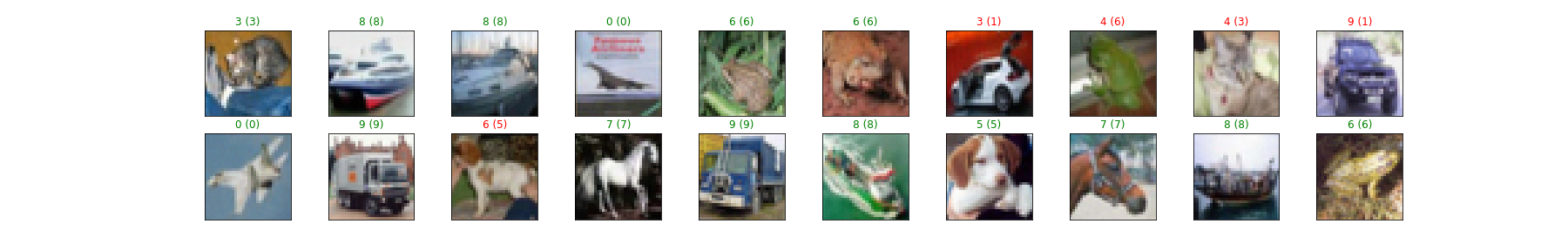}
    \caption{$P_2$ adversarial examples}
  \end{subfigure}

\vspace{-0.2\baselineskip}
  \begin{subfigure}[t]{\textwidth}
    \centering
    \includegraphics[width=0.85\linewidth]{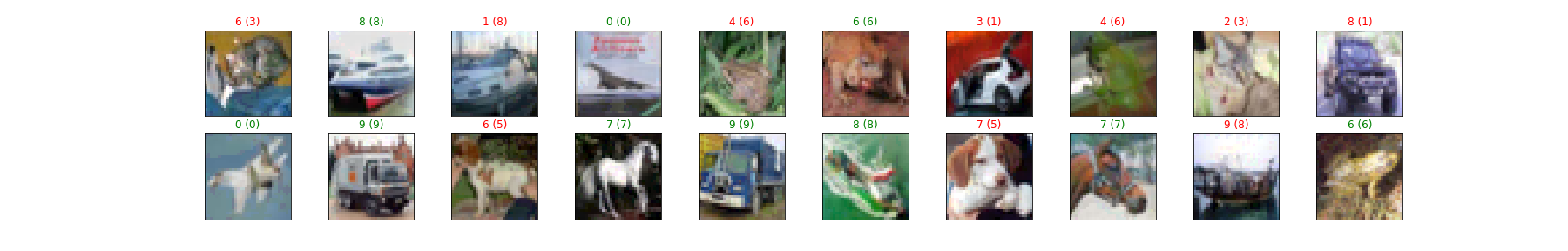}
    \caption{$P_\infty$ adversarial examples}
  \end{subfigure}
  \hfill
  \begin{subfigure}[t]{\textwidth}
    \centering
    \includegraphics[width=0.85\linewidth]{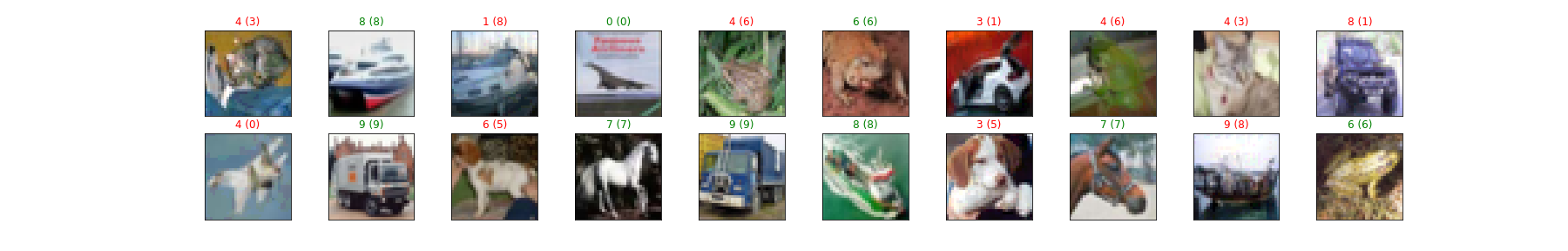}
    \caption{AutoAttack$_\infty$ adversarial examples}
  \end{subfigure}

\vspace{-0.2\baselineskip}
  \begin{subfigure}[t]{\textwidth}
    \centering
    \includegraphics[width=0.85\linewidth]{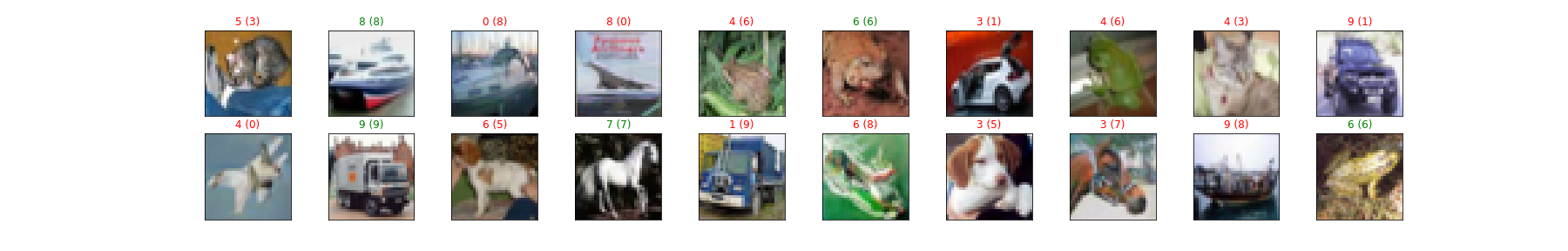}
    \caption{CW$_2$ adversarial examples}
  \end{subfigure}
  \hfill
  \begin{subfigure}[t]{\textwidth}
    \centering
    \includegraphics[width=0.85\linewidth]{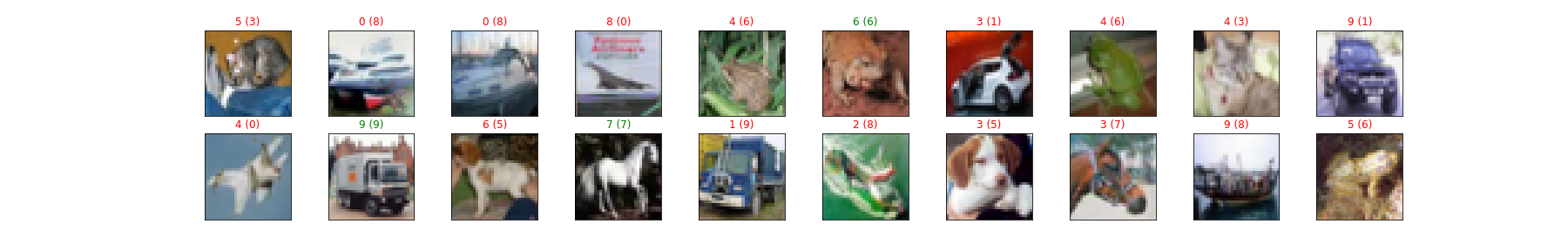}
    \caption{CW$_2^\bullet$ adversarial examples}
  \end{subfigure}

\vspace{-0.2\baselineskip}
  \begin{subfigure}[t]{\textwidth}
    \centering
    \includegraphics[width=0.85\linewidth]{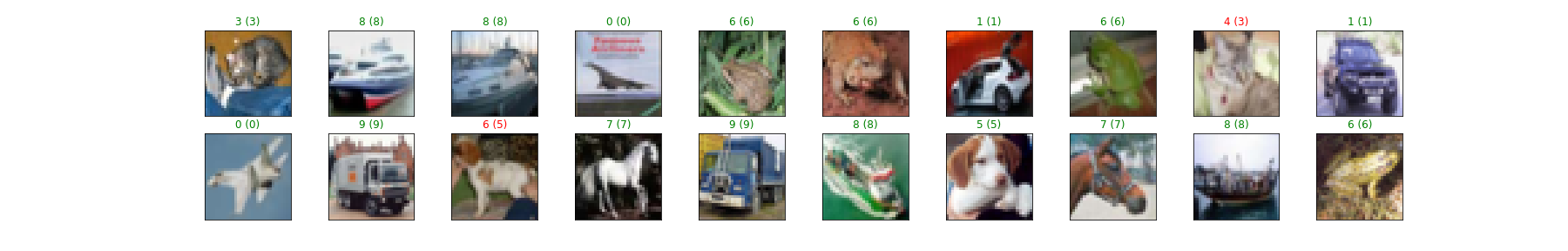}
    \caption{No attack (clean) adversarial examples}
  \end{subfigure}
  \hfill
  \begin{subfigure}[t]{\textwidth}
    \centering
    \includegraphics[width=0.85\linewidth]{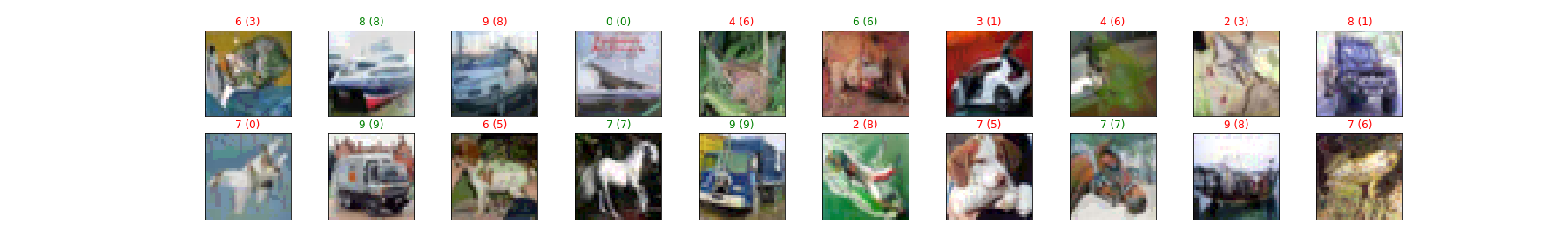}
    \caption{$P_\infty^\bullet$ adversarial examples}
  \end{subfigure}
  \caption{Adversarial examples generated from the hyperparameter-optimised Avg+REx model, for each attack. The predicted class and in parentheses, the true class, are displayed above each image.}\label{fig:adv_examples_Avg_REx}
\end{figure*}

\begin{itemize}
    \item In preliminary experiments without hyperparameter tuning, REx did not benefit a model pretrained solely on $P_\infty$. 
    \item In preliminary experiments without hyperparameter tuning, REx (and the Avg baselines) incurred significant losses in robustness when attempting to train on only one type of perturbation per sample in the batch (in the sense that each sample from a minibatch would only contribute to a single random domain among the seen domains, instead of all seen domains for each sample).
    \item We attempted to just add the variance penalty term to the MSD baseline while still training on the MSD attack. More explicitly, the ERM term consisted in the loss on the MSD attack, and the variance term separately computed $P_1, P_2, P_\infty$ perturbations. This leads to iterations that are twice as expensive, for no observed benefit. Therefore, we instead prefered to define MSD+REx as performing REx+Avg$_\textit{PGDs}$ on a model pretrained with MSD.
    \item On CIFAR10, training on $\{P_\infty, DF_\infty, \textit{CW}_2\}$ is about 8 times more computationally expensive than training on $\textit{PGDs}$ or MSD with 10 iterations per $P_p$ attack (factoring in that in all cases, we validate on all domains every 5 epochs). Since the former leads to a significant advantage in robustness over the ensembles of attacks evaluated here, there is a strong trade-off between computational cost and adversarial robustness when training on those attacks.
\end{itemize}
In figures \ref{fig:adv_examples_Avg} and \ref{fig:adv_examples_Avg_REx}, we show the domains generated for the Avg and Avg+REx models from different attacks, along with the unperturbed data. Above each image is the class predicted by the model, and in parentheses the true class. The classes match the following numbers:

\begin{itemize}
    \item airplane : 0
    \item automobile : 1
    \item bird : 2
    \item cat : 3
    \item deer : 4
    \item dog : 5
    \item frog : 6
    \item horse : 7
    \item ship : 8
    \item truck : 9
\end{itemize}

This highlights how in general, these adversarial examples are not difficult to classify for humans. \textbf{This however also illustrates how $\textit{CW}_2$ perturbations, in spite of being unbounded, tend to be much less perceptible than those stemming from most commonly used bounded attacks, such as $P_\infty$ or $\textit{AA}_\infty$, at tunings where they have the highest attack success rate by far among attacks considered.}

\newpage
\vspace{-0.2\baselineskip}
\vspace{-0.2\baselineskip}
\vspace{-0.2\baselineskip}
\vspace{-0.2\baselineskip}

\end{document}